\documentclass[journal]{IEEEtran}

\usepackage{cite}

\ifCLASSINFOpdf

\else

\fi

\usepackage{epsfig}
\usepackage{graphicx}
\usepackage{amsmath}
\usepackage{amssymb}
\usepackage{array}
\usepackage{tablefootnote}
\usepackage{subfigure}
\usepackage{float}
\usepackage{caption}
\usepackage{setspace}
\usepackage{hyperref}
\hypersetup{hypertex=true,
	colorlinks=true,
	linkcolor=blue,
	anchorcolor=blue,
	citecolor=blue}
\usepackage{algorithm}
\usepackage{algpseudocode}
\usepackage{graphics}

\usepackage{pifont}

\usepackage{multirow}

\usepackage{booktabs}
\usepackage{threeparttable}

\usepackage{amsfonts,amssymb} 

\usepackage[square,sort&compress,numbers]{natbib}

\usepackage{amsmath}

\usepackage{epstopdf}

\usepackage{makecell}

\allowdisplaybreaks

\begin{document}


\title{A Survey on Open-Set Image Recognition}

\author{Jiayin Sun and Qiulei Dong
	\thanks{The corresponding author is Quilei Dong.
		
	Jiayin Sun and Qiulei Dong are with the State Key Laboratory of Multimodal Artificial Intelligence Systems, Institute of Automation, Chinese Academy of Sciences, Beijing 100190, China, the School of Artificial Intelligence, University of Chinese Academy of Sciences, Beijing 100049, China, and the Center for Excellence in Brain Science and Intelligence Technology, Chinese Academy of Sciences, Beijing 100190, China (e-mail: jiayin.sun@nlpr.ia.ac.cn; qldong@nlpr.ia.ac.cn).
	}
}

\markboth{IEEE TRANSACTIONS ON XXX,~Vol.~XX,~2023}%
{Shell \MakeLowercase{\textit{et al.}}: Bare Demo of IEEEtran.cls for IEEE Journals}

\maketitle

\begin{abstract}
Open-set image recognition (OSR) aims to both classify known-class samples and identify unknown-class samples in the testing set, which supports robust classifiers in many realistic applications, such as autonomous driving, medical diagnosis, security monitoring, etc. In recent years, open-set recognition methods have achieved more and more attention, since it is usually difficult to obtain holistic information about the open world for model training. In this paper, we aim to summarize the up-to-date development of recent OSR methods, considering their rapid development in recent two or three years. Specifically, we firstly introduce a new taxonomy, under which we comprehensively review the existing DNN-based OSR methods. Then, we compare the performances of some typical and state-of-the-art OSR methods on both coarse-grained datasets and fine-grained datasets under both standard-dataset setting and cross-dataset setting, and further give the analysis of the comparison. Finally, we discuss some open issues and possible future directions in this community.
\end{abstract}


\section{Introduction}    \label{section:introduction}

\begin{figure}[t]
	\begin{center}
		\setlength{\abovecaptionskip}{0.cm}
		\includegraphics[height=5cm,width=8.5cm]{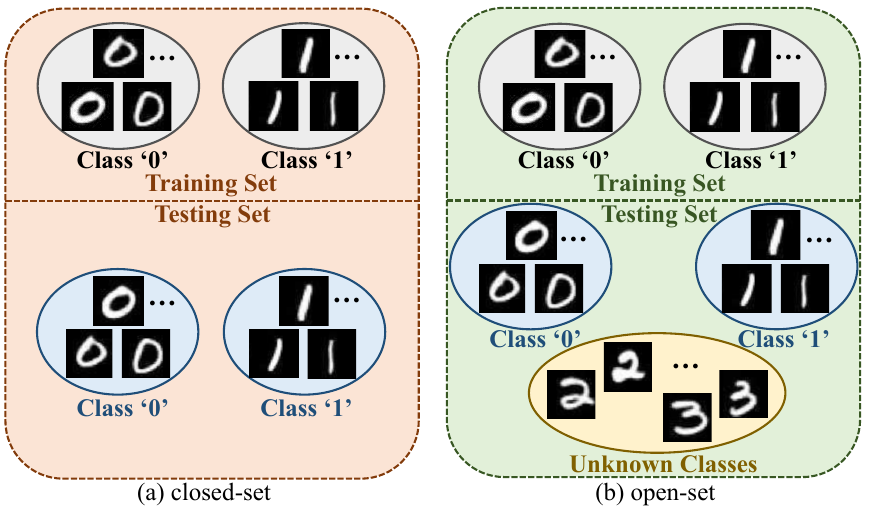}
	\end{center}
	\caption{Comparison between closed-set recognition and open-set recognition (taking the MNIST dataset \cite{MNIST} as an example): A closed-set recognition model only needs to classify such testing images that have the same labels as the training images (\emph{e.g.}, images of the classes `0' and `1'), while an open-set recognition model needs to both classify known-class images and identify unknown-class images in the testing set whose labels do not belong to the training classes (\emph{e.g.}, the classes `2' and `3').}
	\label{fig: difference}
\end{figure}

\IEEEPARstart{T}{he} closed-set image recognition task has achieved a significant breakthrough \cite{VGG, AlexNet, Inception1, Inception23, Inception4, ResNet1, ResNet2, Wide-ResNet, ResNext, DenseNet, ViT, T2T-ViT, PVT, TIT, Swin} due to the development of deep learning techniques in recent years. However, in many real scenarios, there generally exist some new objects whose classes are different from the known training object classes. The existing closed-set recognition methods could not deal with such scenarios effectively, since they would inevitably predict an unknown-class object image as one of the known classes. This issue encourages researchers to focus on the open-set recognition technique, which aims to both classify known-class images and identify unknown-class images. The difference between closed-set recognition and open-set recognition is illustrated in Fig. \ref{fig: difference}.

As indicated in \cite{shift1, shift2, shift3, OSR_survey1}, the distribution shift can be divided into two categories: (i) semantic shift where the labels in the training set and testing set are different, and (ii) covariate shift where the feature distributions (such as image styles) are different between the training samples and testing sample. The main challenge exists in the OSR task is the semantic shift between the training set which only contains known-class samples and the testing set which contains both known-class and unknown-class samples. Since DNNs (Deep Neural Networks) are data-driven models and depend heavily on the identical-distribution assumption, such semantic shift issue would lead the model to predict an unknown-class testing sample as one of the known classes with a high confidence.

To address the above issue, a lot of OSR methods have been proposed in literature, particularly with the development of deep learning techniques. It has to be pointed out that to the best of the authors' knowledge, there have been a few surveys on the OSR task, which were published before or around 2021. However, since the OSR technique has developed very fast due to its adaptation to realistic scenarios and various new DNN-based methods have been proposed in recent two or three years, it would be helpful for researchers in this community to summarize the up-to-date development of this technique. Hence in this paper, a new taxonomy of the existing DNN-based OSR methods is introduced. Then accordingly, we provide a comprehensive review on the recent OSR works, make a comparison of their performances. Moreover, we give a discussion on some open issues and possible future directions in this community. 



\begin{figure*}[t]
	\begin{center}
		\setlength{\abovecaptionskip}{0.cm}
		\includegraphics[height=13cm,width=17cm]{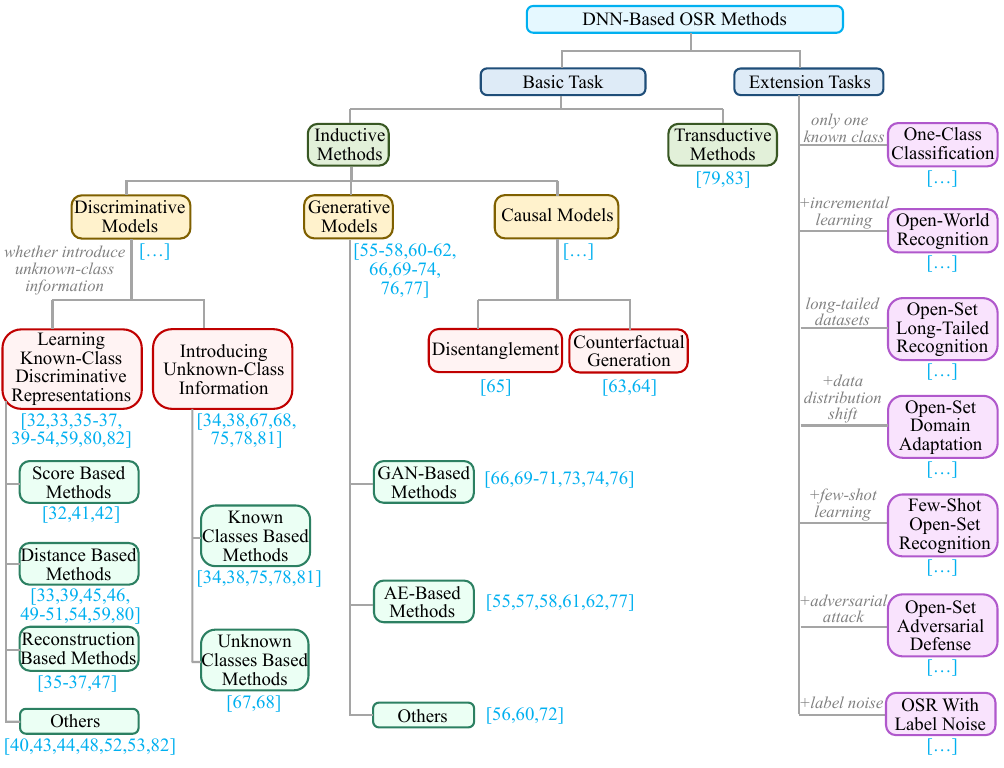}
	\end{center}
	\caption{Taxonomy of the existing DNN-based open-set image recognition methods.}
	\label{fig: global}
\end{figure*}

The main contributions of this paper are as follows: 

\begin{enumerate}
 
	\item[-] We categorize and review the up-to-date DNN-based OSR methods comprehensively, which provides both basic techniques and state-of-the-art processing approaches in this community.
	
	\item[-] We compare and analyze the model performances of the typical and state-of-the-art OSR methods on multiple datasets under the two dataset settings, for the readers' convenience of understanding the common characteristics of the existing OSR methods.
	
	\item[-] We provide some insights into the open issues in existing OSR methods, as well as some future research directions for handling the OSR task.
	
\end{enumerate}

The remainder of this paper is organized as follows. Firstly, we propose a new taxonomy of the existing DNN-based OSR methods in Sec. \ref{section: taxonomy}. Then, we describe the datasets and metrics that are commonly used in the OSR task, and comparison results of some representative DNN-based OSR methods in Sec. \ref{section: benchmarks}. Next, we present some open issues and future research directions for OSR in Sec. \ref{section: future}. Lastly, we end up with conclusion in Sec. \ref{section: conclusion}.

\section{Methodologies}    \label{section: taxonomy}

In this section, we firstly describe the taxonomy utilized to category the existing DNN-based OSR methods, as shown in Fig. \ref{fig: global}. Next, we review the OSR methods in the basic task based on the taxonomy. These methods could be roughly divided into two groups: the inductive methods and the transductive methods, which are introduced in detail, respectively. Finally, we also introduce several extension tasks, where we review some typical methods in the corresponding task.

\subsection{Taxonomy}

According to the original definition, the OSR task could be divided into two groups: the basic OSR task where the models are only required to identify both known and unknown classes in the testing set (\emph{i.e.}, both to conduct multi-class classification on the known-class testing samples and to distinguish the unknown-class testing samples from the known-class ones), as well as some extension tasks where the models need to not only satisfy the requirements of the basic OSR task, but also deal with various situations in real scenarios.

According to whether the testing samples are used in model training, the existing DNN-based methods for handling the basic OSR task could be generally divided into two groups: the inductive methods and the transductive methods, as shown in Fig. \ref{fig: global}. The two groups of methods will be reviewed in the following subsections.

\subsection{Inductive Methods}

Inductive methods consider the testing data to be unavailable at the training stage. Most of the existing OSR methods are inductive methods, which can be further divided into three categories according to the difference of their mainly used models: 1) discriminative models which learn decision rules directly \cite{OpenMax, CAC, PROSER, CROSR, C2AE, GDFR, RPL, Geometric, OVRN_CD, MLOSR, IIOSR, PMAL, CSSR_OSR, OSR_transformer2, MMF_OSR, SCL_OSR, ConOSR, DivAR, STAN_OSFGR, MSD_Net, Gradient_OSR, SLCP, Background_class, OSCR, OPG_OSR, Intra_class, PCC_OSR, OpenAUC, OpenMix, OSR_land1, ODL, IEEE-Access, good-closed-set, GCPL, CPN}, 2) generative models which learn the distributions of the training data \cite{GMVAE-OSR, MGPL, OpenHybrid, CGDL, Capsule, M2IOSR, CPGM, MoEP-AE, OpenGAN, ARPL, G-openmax, OSRCI, AOSR, GAN_MDFM, Difficulty, AKPF, DPLM}, and 3) causal models which introduce causalities into the statistical models which are lazily learned by DNNs \cite{GCM_CF, GCM_CF_SAR, iCausalOSR}.

\subsubsection{Discriminative Models}

Discriminative OSR models \cite{OpenMax, CAC, PROSER, CROSR, C2AE, GDFR, RPL, Geometric, OVRN_CD, MLOSR, IIOSR, PMAL, CSSR_OSR, OSR_transformer2, MMF_OSR, SCL_OSR, ConOSR, DivAR, STAN_OSFGR, MSD_Net, Gradient_OSR, SLCP, Background_class, OSCR, OPG_OSR, Intra_class, PCC_OSR, OpenAUC, OpenMix, OSR_land1, ODL, IEEE-Access, good-closed-set, GCPL, CPN} directly learn discriminative feature representations or classifiers for recognizing both known-class samples and unknown-class samples. According to whether to introduce unknown-class information into the training procedure, the discriminative models can be further divided into two groups: (i) one group of models \cite{OpenMax, IEEE-Access, good-closed-set, CAC, SCL_OSR, ConOSR, DivAR, IIOSR, PMAL, MMF_OSR,  SLCP, PCC_OSR, CROSR, C2AE, GDFR, CSSR_OSR, Geometric, OVRN_CD, MLOSR, OSR_transformer2, STAN_OSFGR, MSD_Net, Gradient_OSR, OSR_land1, ODL, GCPL, CPN} that aim at learning known-class discriminative representations only depending on known-class training samples, and (ii) the other group of models \cite{PROSER, RPL, OPG_OSR, Intra_class, OpenAUC, OpenMix, Background_class, OSCR} that aim at introducing unknown-class information for reducing the discrepancy between the training classes and testing classes.

(i) As for \textbf{the first group that learns known-class discriminative representations}, the existing DNN-based OSR methods can be roughly divided into four types: score-based methods \cite{OpenMax, IEEE-Access, good-closed-set}, distance-based methods \cite{IIOSR, MMF_OSR, GCPL, CPN, CAC, SLCP, PMAL, SCL_OSR, ConOSR, DivAR, PCC_OSR}, reconstruction-based methods \cite{CROSR, C2AE, CSSR_OSR, GDFR}, and others \cite{Geometric, OSR_transformer2, OVRN_CD, MLOSR, Gradient_OSR, OSR_land1, ODL, STAN_OSFGR, MSD_Net}.

\textbf{Score-Based Methods.} DNNs are more vulnerable to the unknown-class samples than traditional methods in the OSR task because of the closed-set assumption of the softmax layer, which is commonly used to obtain classification probabilities as indicating scores. The scores of each sample belonging to all known classes outputted from the softmax layer are sum to 1, and the traditional classification methods take the index where the maximum score occurs as the predicted label, thus taking no account of unknown classes that are left out of the known classes. 

To solve this problem, Bendale and Boult \cite{OpenMax} introduced the Extreme Value Theory (EVT) into modeling the distribution of the distances between the activation vectors of training samples outputted from the penultimate layer of the network and the mean activation vector of each known class. Specifically, the aforementioned distances of each known class were firstly calculated, partial distances which had the most largest values were selected to fit the Weibull distribution as the extreme value distribution of the corresponding known class. At the testing stage, the probability of a testing sample belonging to each known class was calculated according to the distance between the activation vector and the mean activation vector of each known class falling into the modeled extreme value distribution of the corresponding class, thus obtaining the known-class scores. Besides, they also calculated the unknown-class score based on the weighted combination of known-class scores and the rectified weights. Finally, the testing sample is identified as one of the known classes if the maximum score occurred at the known-class scores and was larger than a threshold; otherwise, it would be classified as the unknown classes. The Weibull distribution is commonly used to model the extreme value distributions, and this method paves an early way for later OSR methods to fit the distribution of known-class features and use threshold to distinguish unknown-class samples from known-class samples.

However, due to the calculation complexity, EVT-based scores are gradually abandoned in many OSR methods, and the simple softmax scores combined with a comparing threshold are still widely used. Recently, Dai \emph{et al.} \cite{IEEE-Access} found that the original logit vectors perform better than the probabilistic scores additionally calculated by the softmax layer in the OSR task, since the translation invariance for the input logit vectors weakens the ability of the scores for capturing fine-grained information. Subsequently, Vaze \emph{et al.} \cite{good-closed-set} also emphasized the effectiveness of the logit-based scores for boosting the model performance.

The EVT-based scores can also be regarded as one of the distance-based scores since the scores are calculated based on the distances between the instance-specific features and the class-specific prototypes. Such distance-based scores can be used for identifying different classes because the models are trained to shrink the intra-class distance and enlarge the inter-class distance, which will be described in detail in the following paragraphs.

\textbf{Distance-Based Methods.} Similar to the developing process of traditional closed-set classification methods, a research branch is extended from the study of the classification loss functions, which studies the distance-based loss functions for imposing constraints on the features for learning more compact and discriminative features. This research branch is reasonable for OSR because one of the main reasons that cause the recognition difficulty on unknown-class samples is that the known-class features over-occupy the space that should be reserved for unknown-class features, and cause the confusion between known-class and unknown-class features, as indicated in \cite{OSR_survey5}.

The distance-based loss functions are inspired by Fisher's criterion, which aims at minimizing intra-class variance and maximizing inter-class difference. Such idea that constrains feature representations has now formed a research direction, representation learning. Hassen and Chan \cite{IIOSR} applied a simple representation learning method for handling the OSR task. They took the logit vectors as the feature representations of the input images in a different space (\emph{i.e.}, the logit space), and updated the mean vector of each known class by each training batch. Then, an instance-specific loss term that constrained the $L_2$ distance (\emph{i.e.}, the Euclidean distance) between each training sample and the corresponding class-specific mean vector to be smaller, as well as a pair-wise loss term that constrained the Euclidean distance between different class-specific mean vectors to be larger, were combined with the cross-entropy classification loss (which was the commonly-used loss in the traditional closed-set classification task) for training the network. Jia and Chan \cite{MMF_OSR} added a loss extension to a representation loss, which extracted activation vectors outputted from the penultimate layer of the network and formed a representation matrix for emphasizing the features that had the largest and the features that had the smallest magnitudes at the training stage, thus learning more discriminative feature representations. 

Since the distance-based OSR methods emphasize the feature compactness within each known class as well as the feature discrimination between different known classes, the anchors/prototypes of these known classes are the key to the model performance. Yang \emph{et al.} \cite{GCPL} proposed a convolutional prototype learning system, which utilized a CNN for extracting image features. To boost the intra-class compactness, the known-class features were constrained by a prototype loss to be close to class-discriminative prototypes, which were also learnable during the training stage. In \cite{CPN}, they added a one-versus-all loss for further boosting the feature discriminability, and analyzed the influence of both different unknown rejection criteria and different discriminative losses on the model performance under different open-set recognition deployments. Miller \emph{et al.} \cite{CAC} constrained known-class logit vectors to be clustered at the anchors of the corresponding classes, where each known-class anchor was pre-set at its class coordinate axis thus being equal to a scaled one-hot vector. Compared with general prototype-clustering methods where class anchors were learnable without additional constraints, this method where the anchors were fixed achieved better performance in the OSR task, and the constraints in the logit space reduced unnecessary noises caused by visual variations within each known class.

However, fixed prototypes play a limited role in constraining the feature discrimination. Hence, Xia \emph{et al.} \cite{SLCP} proposed a constraint loss term for controlling the spatial location of these feature prototypes to be more discriminative. Compared with previous methods whose known-class features tended to occupy the same center part of the feature space as the unknown-class features did, this method limited the known-class prototypes to be located at the edge region of the feature space, thus alleviating the confusion between known-class features and the unknown-class features. This constraint was performed by constraining the variance of the distances between the prototypes and the center of the feature space. Considering that there was single prototype representing for each known class in previous methods which ignored the feature diversity within each class, Lu \emph{et al.} \cite{PMAL} designed a prototype mining strategy before optimizing the feature space, which mined high-quality and diverse prototypes for each known class. 

As an effective tool for constraining the distances between pairs of features that are from the same image or different images via data augmentation in the self-supervised tasks, contrastive loss has received more and more attention in recent years. Kodama \emph{et al.} \cite{SCL_OSR} applied a supervised contrastive loss for constraining the feature pairs from the same known class or different known classes, which was convenient for learning latent features that were with intra-class compactness and inter-class discrepancy. Similarly, Xu \emph{et al.} \cite{ConOSR} also utilized supervised contrastive learning for improving the quality of the learned feature representations.

Additionally, some OSR methods aim at designing feature representations or classifiers with angles for improving inter-class similarity and inter-class difference. Park \emph{et al.} \cite{DivAR} proposed to learn divergent angular representations, which improved the global directional feature variation. Cevikalp and Saglamlar \cite{PCC_OSR} introduced a quasi-linear polyhedral conic classifier, which constrained the known-class regions to be L1 or L2 balls.

\textbf{Reconstruction-Based Methods.} There is a view exists in the OSR task that reconstruction is valuable for the model discriminability since known-class samples generally have smaller reconstruction errors than unknown-class samples which the model has never seen. Following this point, some OSR methods make use of the latent feaures in a reconstruction network, reconstructed samples, or reconstruction errors for improving the model discriminability. 

Yoshihashi \emph{et al.} \cite{CROSR} used the latent feature representations in a reconstruction network in addition to the network predictions for classification. They proposed a special autoencoder, deep hierarchical reconstruction network, for extracting latent features at each layer. At the training stage, the network was trained by jointly using the classification loss and the reconstruction loss, where the classification loss used the EVT-based scores \cite{OpenMax}. At the testing stage, the classification scores were used for identifying known- and unknown-class samples. 

Oza and Patel \cite{C2AE} split the whole training process of a autoencoder into two steps. Firstly, they trained the encoder and a classifier which was connected to the latent features by traditional cross-entropy classification loss. Then, the weights of the encoder and the classifier were fixed, and the decoder was trained by a well-designed pair-wise reconstruction loss. At this step, the feature maps input to the decoder were controlled by two parts: the encoded features and conditional vectors via linear modulations, thus obtaining pairs of original/reconstructed images. The decoder was trained to maximize the reconstruction errors of label-unmatched pairs while minimize the reconstruction errors of label-matched pairs. At the testing stage, the encoded feature of the testing image was linearly modulated by all-class conditional vectors, and the corresponding reconstruction errors of all known classes could be obtained, which were compared with a threshold for determining the predicted label.

Huang \emph{et al.} \cite{CSSR_OSR} integrated prototype learning and reconstruction, they proposed to reconstruct class-specific semantic feature maps rather than instance-specific images for boosting the semantic discriminability of the model. They modeled an autoencoder for each known class in the latent space, which was used to reconstruct the feature maps extracted by the backbone encoder from the input images. The class-specific pixel-wise reconstruction error maps were taken as logits and converted to traditional logit vectors by a softmax layer and pooling layer. At the training stage, the model is trained by a classification loss constraining the logit vectors. At the testing stage, the reconstruction errors corresponding to the all-class autoencoders were used for identification.

Perera \emph{et al.} \cite{GDFR} found a new path to make use of reconstruction information, where reconstructed images were used. They firstly trained a generative model (\emph{e.g.} vanilla autoencoder, conditional autoencoder, etc.) for obtaining the reconstructed images for known classes, then they extended the original images by taking the reconstructed ones as new dimensions for learning a classifier.

\textbf{Others.} Some methods aim at mining discriminative information from data augmentation, collective decision, multi-task learning, gradients, space transformation, hierarchical attentions, and even with the help of visual psychophysics. 

Perera and Patel \cite{Geometric} found that extreme geometric transformations could cause different feature representations, hence, they passed the features corresponding to the transformed images through parallel network branches and used majority voting for final prediction. 

With the development of network backbones, vision transformers have been more and more widely applied due to their better performance. Azizmalayeri and Rohban \cite{OSR_transformer2} empirically observed that taking a vision transformer as the backbone and using the softmax probabilities for classification could achieve better performance than other OSR methods. Besides, they also found that delicately selected data augmentations rather than standard training augmentations were helpful for boosting the model discriminability. 

Besides, Jang and Kim \cite{OVRN_CD} proposed to integrate multiple one-vs-rest networks as the feature extractor, and combined the multiple decisions for modeling the final decision score for the input image. Oza and Patel \cite{MLOSR} proposed a multi-task learning method for OSR, which simultaneously trained a classification loss in the latent feature space and a reconstruction loss at the end of an autoencoder, which was similar to \cite{CROSR}. Lee and AlRegib \cite{Gradient_OSR} utilized gradient-based feature representations for recognition, since gradients contained information about how much the model need to be updated for properly representing known-class samples. Baghbaderani \emph{et al.} \cite{OSR_land1} proposed to utilized the transformations among three spaces (\emph{i.e.}, the original image space, the latent feature space, and an abundance space) for exploiting more discriminative information. Liu \emph{et al.} \cite{ODL} proposed an orientational optimization strategy for constraining the feature space and a hierarchical spatial attention mechanism for capturing the global dependencies in the feature space, which further improved the feature discriminability. Sun \emph{et al.} \cite{STAN_OSFGR} proposed an hierarchical attention network for learning fine-grained known-class features, which progressively aggregated both the hierarchical attentioned features and the contextual features in each hierarchical attentioned feature map. Huang \emph{et al.} \cite{MSD_Net} were inspired by the fact that the OSR task is challenging for computer vision algorithms while is easy for human, they incorporated a psychophysical loss and the corresponding network architecture into deep learning, which could support reaction time measurements to simulate human perception.

(ii) As for \textbf{the second group that introduces unknown-class information into model training}, the existing DNN-based OSR methods can be roughly divided into two types according to different sources of the unknown-class information: methods that exploit unknown-class information from known-class samples \cite{PROSER, OpenAUC, OpenMix, OPG_OSR, Intra_class, RPL}, and methods that introduce unknown-class information from outlier-class samples \cite{OSCR, Background_class}.

\textbf{Known-Class Exploitation.} The first group of OSR methods fall into a bottleneck since only aiming at learning more discriminative known-class features or classifiers is not enough to handle the unknown classes that the model will encounter in the testing set. Addressing this problem, some methods aim at exploiting unknown-class information based on known-class images or features via mixup \cite{mixup}, augmentation, splitting, etc. 

Mixup is a data augmentation strategy, which linearly mixes images and the corresponding labels from two different classes. Vanilla mixup \cite{mixup} produces new samples in the input space by using linear interpolations, while manifold mixup \cite{mixup2} produces new features in the latent feature space. Zhou \emph{et al.} \cite{PROSER} learned data placeholders for unknown classes by producing new features via manifold mixup, which were taken as an additional class and used to train the model along with the known-class samples. By generating unknown-class features, the known-class features were constrained to be more compact and discriminative, thus alleviating the over-occupying problem for the OSR task. Besides, they also proposed to learn classifier placeholders, which represented a class-specific threshold for distinguishing unknown-class samples from known-class ones. Specifically, they gave an extra dimension for unknown classes in the output vector of the classifier as the learnable threshold for discriminating between known classes and unknown classes, which was conducted by constraining the value of the extra dimension to be the second largest among all of the dimensions, since the learnable threshold that could distinguish the target class from non-targeted classes was considered to have the ability of distinguishing known classes from unknown classes. 

Besides, Wang \emph{et al.} \cite{OpenAUC} proposed a new metric that couples both closed-set and open-set performance by pair-wise formulation about known-class and unknown-class features, and transformed this formulation into an optimization problem that minimized the corresponding risk. Similar to \cite{PROSER}, they generated unknown-class features by manifold mixup, while the known-class features could be extracted from the training images. Jiang \emph{et al.} \cite{OpenMix} generated high-quality negative images by mixing them, which was proved to reduce both closed space structural risk and open space risk.

In addition to mixup, some methods use augmentation-based similarity learning, intra-class splitting, or multi-class interaction for mining unknown-class information based on known-class data.

Esmaeilpour \emph{et al.} \cite{OPG_OSR} adopted similarity loss for encouraging the model to learn how to discriminate between known-class and unknown-class samples, where unknown-class images were generated by distribution-shifting data augmentation. Specifically, they conducted two steps for the model training. At the first step, unknown-class images were generated via randomly rotating original images by multiples of $90^\circ$. At the second step, a classification model was trained with both known-class training images and generated unknown-class images, and two losses are imposed: a cross-entropy loss for classifying known-class images and a binary cross entropy loss for learning clusters for both known-class and unknown-class images via similarity supervision.

Schlachter \emph{et al.} \cite{Intra_class} exploited unknown-class information from known-class training images. Specifically, they split training images into typical subsets and atypical subsets by a closed-set classifier. Then, the atypical subsets were served as unknown-class data and were added into the training set as the ($C+1$)-th (where $C$ is the number of known classes) class for training a ($C+1$)-class classification model. These two methods are relatively straightforward and simple for building unknown-class data. To capture more precise unknown-class information, some researchers explore learnable strategies for learning unknown-class features based on known-class features. 

Chen \emph{et al.} \cite{RPL} defined reciprocal points for capturing unknown-class information for each known class, containing both extracted image features from other known classes and a set of learnable features served as unknown-class features. During training, the extracted features from one known class were pushed to be far from the reciprocal points of the corresponding known class, by which the learned known-class features were located in the periphery of the feature space, while the unknown-class features were limited within a bounded region at the center of the feature space. In this case, the unknown space was shrunk and bounded, which could prevent the classification network producing high confidence for unknown-class test samples.

\textbf{Outlier-Class Introduction.} The methods that exploit unknown-class information from known classes are still limited by the training set. Sometimes, when the distribution of the training set significantly deviates from the distribution of the testing set, or when the data volume is small, the original training set is insufficient to support the model in exploring effective unknown-class information. In this case, some methods try to introduce outlier data into the model training. 

The usage of outlier data can be traced to the object detection task, where the classification network is also trained by a background class indicating that there is no object of interest in the proposal. The introduction of outlier-class samples can prevent the network from outputting over-confident erroneous prediction for unknown-class samples. Dhamija \emph{et al.} \cite{OSCR} utilized this idea for reference, and they took the digital images from some classes of the NIST Letters dataset \cite{NIST_Letters} as the known-class samples, while taking the rest-class images from NIST Letters as the unknown-class samples for testing and taking the images from CIFAR10 \cite{CIFAR10} and MNIST \cite{MNIST} as the outlier samples. They firstly found that unknown-class features usually presented lower feature magnitude as well as higher SoftMax entropy than known-class features. Based on this default observation, they designed an entropy-based loss and a magnitude-based loss to constrain both the known-class samples and outlier samples in the model training for increasing such separation, which boosted the model robustness to the unknown-class samples that are unavailable in training. 

Such operations that introduce outlier-class samples in the model training were also adopted (called outlier exposure) in the out-of-distribution (OOD) detection task \cite{OOD_outlier1, OOD_outlier2}, where the model ability of rejecting unknown-class samples were more focused on. However, the known-class samples were categorized into one group in the outlier-class related loss, either in the open-set method \cite{OSCR} or in the OOD detection methods \cite{OOD_outlier1, OOD_outlier2}, which could affect the closed-set classification performance in the OSR task. Besides, these methods adopted a same margin between different known classes and outlier classes on different datasets, which limited the open-set recognition performance. 

To address the problems mentioned above, Cho and Choo \cite{Background_class} preferred multiple distance-based classifiers based on the principle of linear discriminant analysis (LDA) rather than the commonly-used SoftMax classifier. Besides, they designed a class-inclusion loss which encouraged each outlier sample to be far away from the nearest class-wise hypersphere classifier, which further boosted the model discriminability. Suggested by \cite{outlier_data}, they chose images from the ImageNet dataset as the outlier-class samples. In addition to the discriminative methods introduced above, a generative OSR method \cite{OpenGAN} also adopts outlier exposure to augment the training set, which will be introduced in the following paragraphs.

\subsubsection{Generative Models}

With the development of generative models, more and more OSR methods pay attention to adopt generative learning techniques for boosting the model discriminability. Generative OSR models \cite{GMVAE-OSR, MGPL, OpenHybrid, CGDL, Capsule, M2IOSR, CPGM, MoEP-AE, OpenGAN, ARPL, G-openmax, OSRCI, AOSR, GAN_MDFM, Difficulty, AKPF, DPLM} mainly learn the distributions from known-class samples, based on which the discrimination criterion of how to both identify unknown-class samples and classify known-class samples is built. According to the specific generative models utilized, these methods can be further divided into three groups: Generative Adversarial Network (GAN)-based methods \cite{OpenGAN, ARPL, G-openmax, OSRCI, GAN_MDFM, Difficulty, AKPF}, Auto-Encoder (AE)-based methods \cite{GMVAE-OSR, MGPL, CGDL, Capsule, CPGM, MoEP-AE, DPLM}, and others \cite{OpenHybrid, M2IOSR, AOSR}.

\textbf{GAN-Based Methods.} Considering that GANs are able to generate various new samples, some methods make use of GANs for generating unknown-class samples to fill the missing information about the open space. They model the data distributions implicitly by adversarial training rather than explicitly by fitting a specific parameterized distribution. Most of GAN-based methods generate unknown-class samples or features only based on the known-class training samples, and seek for different assumptions about the position of the unknown-class samples/features as well as different strategies to make use of the generated unknown-class samples/features. 

Ge \emph{et al.} \cite{G-openmax} assumed that unknown-class samples lay in the mixed distribution of known-class distributions. They modified the training process of the conditional GAN, where several prior known-class distributions were mixed. Based on the generated unknown-class samples, they extended OpenMax \cite{OpenMax} by providing the explicit probability estimation about unknown classes.

Neal \emph{et al.} \cite{OSRCI} assumed that unknown-class samples lay outside of any known-class distribution in the feature space, but simultaneously resembled known-class images in the pixel space. Each unknown-class sample was generated from a known-class training image based on an encoder-decoder-GAN. Specifically, they minimized the reconstruction errors between the input known-class image and the generated image for guaranteeing the similarity in the pixel space, and simultaneously minimized the classification probability of the generated image to be classified into the corresponding known class. Then, the generated images were taken as one additional class to augment the training set and used to train a ($C+1$)-class (where $C$ is the number of the known classes) classifier. 

Jo \emph{et al.} \cite{GAN_MDFM} assumed that the unknown-class features lay in adjacent feature space to the known classes. Hence, they combined the generative in the GAN with a marginal denoising autoencoder so as to model the distribution away from each known class. With the unknown-class features generated, they trained a $C$-class classification model with an additional entropy-based regularization loss for encouraging the model to have high uncertainty on unknown-class features, which tightened the decision boundary of known classes. 

Chen \emph{et al.} \cite{ARPL} extended their previous discriminative OSR method \cite{RPL} by generating confusing samples based on an adversarial training strategy for boosting the model discriminability. Specifically, the generated features were constrained to cheat the discriminator, while they were also constrained to be close to the reciprocal points by maximizing the output entropy from the classifier. At this time, the generated confusing features were located at the boundary between known-class features and reciprocal points. After feature generation, they designed three losses for training the classification model, including two adversarial losses that both maximizing the distance each reciprocal point to its corresponding known-class prototype and limiting the distance within a learnable margin for both reducing the empirical classification risk and bounding the open space risk, and a distance-based entropy loss constraining the generated confusing features to be located near to the reciprocal points for further enhancing the distinguishment between known classes and unknown classes. 

Inspired by \cite{RPL, ARPL} which pushed known-class features to the periphery of the feature space while constrained the unknown-class features within the center of the feature space, Xia \emph{et al.} \cite{AKPF} generated unknown-class features that were located near known-class prototypes while outside a learnable distance to the center of the known-class prototypes. They also introduced adversarial motion properties that enabled the adversarial motion of the distance margin, which further reduced both the empirical risk and the open space risk. 

However, the above methods usually generated unknown-class samples outside the decision boundaries that were easy to be discriminated, and neglected the ``hard negative" samples that mattered more to the open-set recognition performance. To address this problem, Moon \emph{et al.} \cite{Difficulty} took diverse difficulty levels of generated unknown-class samples into consideration from the classifier's perspective. They trained a multi-group convolutional classification network and a copied counterpart whose layers were separated by the same predefined criteria based on multi-level knowledge distillation for generating features in hard- or easy-difficulty level, and concatenated with a GAN for generating multiple features at different difficulty levels. As done in \cite{ARPL}, the generated unknown-class features were given uniform probabilities as labels when finally training the classifier.

Considering that sometimes outlier datasets could be available, some discriminative OSR methods \cite{OSCR, Background_class} have introduced outlier samples for simulating unknown-class samples. However, such models perform poor generalization ability on diverse unknown-class samples, since the outlier samples utilized in training could not cover the open space exhaustively. To address this problem, Kong and Ramanan \cite{OpenGAN} proposed OpenGAN, which trained a GAN combined with a $C$-way classifier based on both the original known-class training samples and the introduced outlier-class samples, and took the discriminator in the GAN for distinguishing unknown-class samples from known-class ones. The model selection was also operated based on outlier validation samples, which was found effective even the outlier validation samples were sparse or biased.

\textbf{AE-Based Methods.} As mentioned before, some reconstruction-based discriminative OSR methods distinguish unknown-class samples from known-class ones based on the discrepant reconstruction errors outputted from the autoencoder. In recent years, some generative OSR methods have utilized autoencoders to model the known-class distributions explicitly, such that unknown-class samples could be rejected if it does not belong to one of the modeled known-class distributions, and known-class samples could also be classified according to which known-class distribution it belongs to. 

As a typical autoencoder, variational autoencoder (VAE) \cite{VAE} has been widely used in many visual tasks, which models the known-class samples as a standard Gaussian distribution. However, using VAE to model feature distribution of all known classes as a Gaussian distribution would damage the discriminability between two different known classes. Sun \emph{et al.} \cite{CGDL} extended the VAE into a class-discriminative autoencoder based on the probabilistic ladder architecture \cite{VAE_ladder}. Specifically, the encoder encoded each image into two distribution parameters in a Gaussian (\emph{i.e.}, mean and standard deviation). The latent feature sampled by the learned distribution was forced to approximate the Gaussian distribution of the corresponding known class, whose standard deviation was an identity matrix and mean was mapped from the one-hot label. In the inference stage, an unknown-class image could be detected not only based on its deviated distribution, but also according to its high reconstruction error. Subsequently, they also extended the adversarial autoencoder (AAE) \cite{AAE} to another class-discriminative autoencoder in a similar manner \cite{CPGM}. 

Inspired by the above method, Guo \emph{et al.} \cite{Capsule} replaced the CNN backbone with the capsule network \cite{capsule_network}, where each latent neuron in the network outputted a vector instead of a scalar, such that the encoded latent features could represent more diverse information. Rather than using an additional classifier for enabling the latent features to be discriminative as done in \cite{CGDL, CPGM}, they utilized a contrastive loss with a margin for forcing the encoded features to be located at the corresponding known-class region and keeping different known-class features to be far from each other. 

However, modeling each known-class-feature distribution as a single Gaussian could not well represent the intra-class discrepancy. To address this problem, Li and Yang \cite{DPLM} assumed that each known-class-feature distribution followed a Gaussian mixture distribution, which could represent the intra-class discrepancy by different Gaussian components. They embedded the neural Gaussian Mixture Model \cite{neural_GMM} into the autoencoder to mapping the latent feature into the marginal distribution, which was called a dual distribution since two opposite probabilities could be derived from it: i) the probability of the latent feature belonging to known classes, and ii) the probability of the latent feature belonging to unknown classes. Either known-class features or unknown-class features encoded by the trained model could form a distribution peak, hence, a testing sample could be recognized according to the distribution peaks in the latent space. Once it lay close to the known-class peak, it would be further classified by the known-class classifier in the latent space. 

Similarly, Cao \emph{et al.} \cite{GMVAE-OSR} also modeled the feature distribution of each known class as a Gaussian Mixture, but they directly modified the Gaussian Mixture VAE \cite{GMVAE} rather than embedding the neural Gaussian Mixture Model. Liu \emph{et al.} \cite{MGPL} combined conditional VAE \cite{CVAE} and prototype learning to constrain the feature distribution of each known class as multiple Gaussians, which could also be regarded as a Gaussian Mixture. Considering that some complex feature distribution could not be represent by Single Gaussian or Gaussian Mixture (\emph{e.g.}, sub-Gaussian and super-Gaussian), Sun \emph{et al.} \cite{MoEP-AE} introduced the mixture of exponential power distribution into the network based on a new re-parameterization strategy, which modeled feature distributions from different known classes by discrepant mixtures of exponential power distributions. 

Generally, the main research trend of this group of OSR methods aim to seek for more effective explicit representation about the known-class feature distributions, as well as investigate how to make better use of the reconstruction errors for boosting the feature discriminability, where the second motivation coincides with that of reconstruction-based discriminative OSR methods.

\textbf{Others.} In addition to the GAN-based and AE-based methods, there also exist some OSR methods that adopt other generative models (\emph{e.g.}, flow-based models) \cite{OpenHybrid}, or only take the encoder for modeling feature distributions \cite{M2IOSR}, or generate instance-weights for the transfer learning \cite{AOSR}.

Flow-based generative models \cite{Flow1, Flow2, Resflow} generate images or features of comparable quality to GANs, but can model the training distributions explicitly due to its invertible architecture, which also model the training-class feature distribution in the latent space as a standard Gaussian as done in VAEs. Zhang \emph{et al.} \cite{OpenHybrid} combined a typical flow network Resflow \cite{Resflow} with a known-class classifier in the latent space. Resflow was used as a density estimator for detecting unknown classes, while the latent classifier was used for maintaining the known-class classification accuracy. 

Considering that image-level reconstruction in AE-based methods would take all image pixels into account where many pixels were irrelated to the class or even easy to cause misleading, Sun \emph{et al.} \cite{M2IOSR} removed the decoder in an autoencoder, and used a Kullback-Leibler (KL) divergence loss to constrain the latent features such that the feature distribution of each known class was modeled as a single Gaussian. Besides, they designed a multi-scaled mutual information maximization strategy for establishing the dependence between the input image and its latent features, which further improved the feature discriminability. 

Another innovative generative OSR method was to learn intermediate vectors used for generating new samples rather than directly generate images or features, which was proposed by Fang \emph{et al.} \cite{AOSR}. Motivated by both the transfer learning theory and probably approximate correct theory, they aligned the known-class samples to the samples in an auxiliary domain (representing unknown-class domain) by learnable instance-weights. Through such instance-weighting strategy, the unknown-class samples can be detected by the instance-weights. 

\subsubsection{Causal Models} No matter the discriminative models or the generative models mentioned above, both of them are easy to fall into the trap of lazy learning, since once a model has sought a group of weight parameters that could minimize the loss function, it would no longer update these weight parameters. Such lazy learning would inevitably lead the model to learn the correlation relationships which are easy to learn but are relatively fragile. To address this problem, some causal models have been proposed \cite{iCausalOSR, GCM_CF, GCM_CF_SAR}, which aim to seek for causal relationships from the training data, which alleviate the confounders in the features learned by the non-causal OSR methods in two manners: disentangling the robust representations from the highly-coupled correlated features \cite{iCausalOSR}, and generating more reliable known-class samples based on counterfactual generation \cite{GCM_CF, GCM_CF_SAR}.

\textbf{Causal Disentanglement.} Yang \cite{iCausalOSR} proposed an invertible causal model for open-set recognition, which consisted of an invertible encoder (i-RevNet \cite{i-RevNet} was used here) for encoding images into features and class functions for providing the distribution prior of the encoded features belonging to each known class. Different from AE-based generative OSR methods that modeled the feature distributions as explicit fixed distributions, the class functions here were built in the form of Structural Causal Models (SCMs) \cite{SCMs} each of which was a directed acyclic graph. At the training stage, the encoded features were constrained to match the corresponding-class prior class functions. At the inference stage, similar to the inference strategy in most of AE-based generative OSR methods, the samples were classified/detected according to their probabilities belonging to these class functions. 

\textbf{Counterfactual Generation.} Since the generative models have the strong ability of modeling the known-class feature distributions and training samples may be insufficient in some cases, generative OSR methods have become the main stream in handling the OSR task. However,  most of these methods generate samples or features based on the one-hot class labels directly, which neglect the non-causal confounders infiltrating into different known classes, leading to the bias of the generated samples or features. To address this problem, some methods adopt counterfactual generation for generating more reliable samples or features \cite{GCM_CF, GCM_CF_SAR}.

Yue \emph{et al.} \cite{GCM_CF} proposed a counterfactual generation method for handling the OSR task based on TF-VAEGAN \cite{TF-VAEGAN}, which consisted of an encoder for encoding the images into the latent features, a decoder/generator for reconstructing/generating counterfactual images based on both the latent features and the provided one-hot labels, and a discriminator for distinguishing the real images (True) from the counterfactual images (False). They trained the network with three losses: a beta-VAE loss for modeling the latent feature distribution as the isotropic Gaussian distribution, a contrastive loss for minimizing (or maximizing) the reconstruction errors between an input image and its reconstructed image based on the matched (or mismatched) labels, and a GAN loss for encouraging the generated image to deceive the discriminator. At the inference stage, similar to \cite{C2AE}, the minimal distance between the testing image and its $C$ counterfactual images generated based on $C$ one-hot known-class labels was compared with a threshold for detecting unknown-class images. If the distance is smaller than the threshold, the testing images was predicted as the known class corresponding to the minimal distance. 

Zhou \emph{et al.} \cite{GCM_CF_SAR} applied the similar method in handling the open-set Synthetic Aperture Radar (SAR) image target recognition task, which still outperformed other OSR methods.

\subsection{Transductive Methods} 

Transductive methods consider the testing samples are available at the training stage, and make use of both labeled training set and unlabeled testing set for model training. As proved in other visual tasks such as zero-/few-shot learning \cite{ZSL-transductive1, ZSL-transductive2, FSL-transductive1, FSL-transductive2} and domain adaptation \cite{DA-transductive1, DA-transductive2, DA-transductive3}, transductive learning can effectively alleviate the distribution shift problem, which has inspired two transductive OSR methods \cite{S2OSC, IT_OSR}. 

Yang \emph{et al.} \cite{S2OSC} was the first transductive OSR method. Firstly, they filtered out some out-of-class samples from the testing samples based on a score-based strategy, which were simultaneously pseudo-labeled by a baseline classification model. Then, the model was updated by jointly making use of the original training samples as well as the filtered out pseudo-labeled testing samples. 

Although such transductive learning method has improved the model performance, two open issues still remain: (i) sample selection issue: how to select testing samples whose pseudo-labels are more reliable; (ii) known/unknown-class imbalance issue: the number of the known-class samples (containing the original training samples and the selected testing samples that are pseudo-labeled as known classes) is usually larger than that of the unknown-class samples (only containing the selected testing samples that are pseudo-labeled as unknown classes). 

To address these two issues, Sun and Dong \cite{IT_OSR} designed a sampling strategy and a generation approach in their proposed iterative transductive OSR framework. Specifically, they designed a dual-space consistent sampling strategy for sample selection, which removed the unreliable testing samples whose pseudo-labels assigned in the output space were inconsistent with most of their neighbors in the feature space from the candidates. Besides, they designed a conditional generation network for feature generation, which added an additional discriminator that distinguished between known-class features and unknown-class features for improving the discriminability of the generated features. Then, an iterative transductive OSR framework was proposed based on their designed sampling and generating approaches, which iteratively employed sample selection, feature generation, and model update.

\subsection{Extension Tasks}

In addition to the basic OSR task, we also introduce some extension tasks. In reality, the OSR task is deployed in a complex and varying environment. Here, we briefly review some representative methods in 7 typical extended scenarios of the OSR task: 1) one-class classification where only one known class is available, 2) open-world recognition where the training data incrementally increases, 3) open-set long-tailed recognition where the data distribution presents a long-tailed distribution, 4) open-set domain adaptation where data distribution also exists, 5) few-shot open-set recognition where the training data is very sufficient, 6) open-set adversarial defense where the input images are adversarially attacked, and 7) open-set recognition with label noise where the labels are noisy or inaccurate. 

\subsubsection{One-Class Classification}

In the common scenario, more than one known class exists in the training set, and training the model to be able to distinguish among different known classes also improves the model ability of distinguishing unknown classes to some extent. In an extreme scenario, only a single known class is available, called one-class classification \cite{OCC_survey1, OCC_survey2, OCC_survey3}, where some discriminative OSR methods that rely on boosting the known-class feature discriminability to improve the OSR performance are likely to fail. 

To address this problem, some discriminative methods embedded the one-class support vector machine into the loss functions \cite{OCC1, OCC2}, or applied specific data transformations for capturing the unique geometric structural information of the known class \cite{OCC3, OCC4}. However, since the negative samples were unavailable in the model training, bias would exist in the learned decision boundary. Considering that generative models could model the distribution of the known-class data, generative methods were widely used in the one-class classification task, which recognized the unknown-class samples based on their deviated distributions from the known-class distribution \cite{OCC5, OCC6}. Besides, the reconstruction errors could be still utilized to distinguish unknown-class samples from known-class samples. 

\subsubsection{Open-World Recognition}

In the common scenario, the dataset is generally static and fixed, and OSR models learn from the existing dataset only at one time. However, data in the realistic scenarios usually presents dynamically, and new data can be obtained periodically or even continuously. It is costly to re-train the model every time. Under such application demand, a series of open-world recognition methods have been proposed, which aim to continuously detect and add new classes encountered \cite{OWR0, OWR2, OWR1}. 

Bendale and Boult \cite{OWR0} firstly proposed the open-world recognition concept, and also extended the nearest class mean classifiers to the open-world recognition task. Cao \emph{et al.} \cite{OWR2} proposed a progressive transductive method, which selected unlabeled new samples and provided them pseudo labels based on clustering results for updating the feature prototypes. Wu \emph{et al.} \cite{OWR1} replied on graph representation and learning for predicting and utilizing new-class samples, where a graph network was used for extrapolate embeddings for features extracted from new data based on a feature-level graph and a prediction network was used for predicting pseudo labels for the new features. Generally, clustering and transductive learning are two common tools for handling the open-world recognition task. The nearest class mean can be considered as a clustering strategy since it pulls the samples to be close to its nearest neighbors. The graph network is also a common tool for transductive learning. With transductive learning, a model can progressively make use of new samples, even if their labels are unavailable. With clustering, the model can be updated with relatively smaller cost.

\subsubsection{Open-Set Long-Tailed Recognition}

Another problem that an OSR model would encounter in the realistic scenarios is that the data has a long-tailed distribution, and the model would prefer the majority classes where the number of samples is dominant while neglect the minority classes whose samples are significantly insufficient. The long-tailed problem is an extreme case of the class imbalance problem. There are some simple strategies for alleviating the model bias towards majority classes, \emph{e.g.}, data re-sampling technique (including down-sampling from the majority classes and over-sampling from the minority classes) and loss re-weighting (increasing/decreasing the loss weights for the minority-/majority-class samples). 

Recently, the open-set long-tailed recognition task has received more and more attention, how to mine effective information in the minority classes in an open-set environment becomes a key problem in this task. Liu \emph{et al.} \cite{OSLT1, OSLT2} firstly formally defined the open-set long-tailed recognition task, and handled it based on a dynamic meta-embedding mechanism. The meta-embedding mechanism associates the majority-class visual features with the minority-class visual features for enabling the model to be robust to the minority classes, and dynamically calibrated the feature norm based on the visual features in the memory bank for supporting open-world recognition. Cai \emph{et al.} \cite{OSLT3} proposed a distribution-sensitive loss, which provided larger weights to minority-class samples when constraining the intra-class distances to be minimized. Besides, they designed a distance-based metric for recognition according to the feature distance to the clusters. 

\subsubsection{Open-Set Domain Adaptation}

As mentioned in Sec. \ref{section:introduction}, the distribution shift contains both semantic shift that exists in the common OSR task and covariate shift. In the open-set domain adaptation task which was firstly proposed in \cite{OSDA1}, both semantic shift and covariate shift exist. In other words, unknown-class samples would exist in the testing set, besides, known-class samples in the training set and testing set are in different domains (\emph{i.e.}, source domain and target domain, respectively). Most of existing closed-set domain adaptation methods aim to align the whole target domain with the source domain according to the labeled known-class samples in the source domain as well as the unlabeled (or partly labeled) samples in the target domain. However, unknown-class samples in the target domain would be improperly aligned with the known-class samples in the source domain in the open-set domain adaptation task, which would harm the distinguishability between known classes and unknown classes. 

To address this problem, Busto and Gall \cite{OSDA1} added an implicit outlier detection mechanism when assigning the images in the target domain to some classes in the source domain, thus the images that didn't belong to known classes could be discard in the assignment. To separate unknown-class samples in the target domain from known-class samples in the target domain, Saito \emph{et al.} \cite{OSDA2} adopted adversarial training between a feature extractor and a ($C+1$)-class classifier, where the classifier was trained to not only classify the known-class source features but also distinguish between known classes and unknown classes according to the probability of the ($C+1$)-th class, while the feature extractor was trained to cheat the classifier. Liu \emph{et al.} \cite{OSDA3} were inspired by the observation that the gap between known-class samples in the target domain and known-class samples in the source domain is much smaller than the gap between unknown-class samples in the target domain and known-class samples in the source domain. They designed a coarse-to-fine weighting mechanism, which iteratively operated two steps: multi-binary classifier training step that measured the similarity of an target image to each source known class, and the binary classifier learning step that trained a binary classifier based on both known-/unknown-class target samples selected by their high/low similarities to the source classes.

\subsubsection{Few-Shot Open-Set Recognition}

The common OSR task is deployed in large-scale datasets. However, in some extreme realistic scenarios, the number of training samples in each known class is extremely small, and such task is called few-shot open-set recognition task \cite{FSOSR1}. 

To handle this task, Liu \emph{et al.} \cite{FSOSR1} extended a closed-set few-shot learning model into the open-set environment. They added some pseudo-unknown samples into the model training via an open-set distance based loss term. Jeong \emph{et al.} \cite{FSOSR2} recognized unknown-class samples based on their larger difference from the transformed prototypes. Wang \emph{et al.} \cite{FSOSR3} proposed an energy-based model, where the samples that deviated from either the class-wise features or the pixel-wise features of the few-show known-class samples were assigned larger energy scores.

\subsubsection{Open-Set Adversarial Defense}

The open-set adversarial defense task combines the open-set recognition task that aims to both classify known classes and identify unknown classes during testing and the adversarial defense task that aims to enable the network to defend against the imperceptibly adversarially-perturbed images, and was firstly proposed in \cite{OSAD1}. 

To handle this task, Shao \emph{et al.} \cite{OSAD1} proposed a open-set defense network consisting of an encoder with denoising layers and a classifier for learning noise-free features. Further, they incorporated a decoder for reconstructing clean images, added a self-supervision loss for improve the feature discriminability, and a clean-adversarial mutual learning mechanism where another classifier (which handled clean images) was mutually learned with the original classifier (which handled adversarial images) for facilitating the feature denoising \cite{OSAD2}.

\subsubsection{OSR With Label Noise}

In the common OSR scenario, the model relies heavily on clean labels. However, providing labels to large-scale dataset is very costly and error-prone, and realistic data inevitably contains noisy/incorrect labels. The OSR with label noise task was firstly proposed in \cite{OSNL1}, following which some researchers have studied in this task.

Wang \emph{et al.} \cite{OSNL1} proposed an iterative learning framework, which iteratively detected noisy labels, enlarged the difference between clean labels and noise labels, and applied a re-weighting module for encouraging the model to learn more from clean labels rather than noisy labels. Sachideva \emph{et al.} \cite{OSNL2} made use of the subjective logic loss \cite{SL_loss}, which could yield higher losses on closed-set noisy samples while lower losses on open-set samples. Unlike the above methods, Wei \emph{et al.}
\cite{OSNL3} empirically demonstrated that open-set noisy labels would even help boost the model robustness against noisy labels, and introduced open-set samples with dynamic noisy labels into the model training as a regularization term.

\section{Datasets, Metrics, and Comparison}    \label{section: benchmarks}

\subsection{Datasets}

In this section, we introduce the common multi-class datasets used in the OSR task, including coarse-grained datasets and fine-grained datasets. In comparison to coarse-grained datasets, the images in fine-grained datasets usually have higher inter-class similarity and lower intra-class similarity. Hence, some detailed processing operations are needed in handling the fine-grained datasets. 

To simulate open-set scenarios, some classes are selected as known classes, while some classes are selected as unknown classes. According to the class sources, the data deployment can be divided into two categories: the standard-dataset setting where known and unknown classes are from a same dataset and the cross-dataset setting where known and unknown classes are from different datasets.

\subsubsection{Coarse-Grained Datasets}

Under the standard-dataset setting, five datasets are used in \cite{CGDL, OpenMax, CAC, CROSR, C2AE, GDFR, RPL, ARPL, OSRCI, OpenHybrid, PROSER, MoEP-AE, IT_OSR, STAN_OSFGR}:
\begin{enumerate}
	\item[-] \textbf{MNIST}: In this dataset, the images are taken from the Mixed National Institute of Standards and Technology (MNIST) datasets \cite{MNIST}, which contains 70000 10-class hand-written digit images (28$\times$28), including 60000 training images and 10000 testing images. 6 classes are randomly selected as the known classes, while the rest 4 classes are taken as the unknown classes. 
	
	\item[-] \textbf{SVHN}: In this dataset, the images are taken from the Street View House Numbers (SVHN) dataset \cite{SVHN}, which contains 99289 10-class street-view house digit numbers (32$\times$32), including 73257 training images and 26032 testing images. Similarly, 6 classes are selected as known classes, while the rest 4 classes are taken as the unknown classes.
	
	\item[-] \textbf{CIFAR10}: In this dataset, the images are taken from the CIFAR10 dataset \cite{CIFAR10}, which contains 60000 10-class natural object images (32$\times$32), including 50000 training images and 10000 testing images. Similarly, 6 classes are selected as known classes, while the rest 4 classes are taken as the unknown classes.
	
	\item[-] \textbf{CIFAR+10/+50}: In this dataset, the images are taken from both the CIFAR10 and CIFAR100 \cite{CIFAR100} datasets. Similar to CIFAR10, CIFAR100 contains 60000 100-class natural object images (32$\times$32), including 50000 training images and 10000 testing images. The 10 known classes are fixed to the 10 classes in the CIFAR10 dataset, and 10 or 50 classes are randomly selected from the CIFAR100 dataset for CIFAR+10 or CIFAR+50 as the unknown classes. 
	
	\item[-] \textbf{TinyImageNet}: In this dataset, the images are taken from the Tiny-ImageNet dataset \cite{TinyImageNet}, which is a 200-class subset of the ImageNet dataset \cite{ImageNet} that contains 120000 natural object images (64$\times$64), including 100000 training images, 10000 evaluating images, and 10000 testing images. 20 classes are selected as known classes, while the rest 180 classes are taken as the unknown classes.
	
\end{enumerate}

Under the cross-dataset setting, the 10-class CIFAR10 dataset is taken as the known-class dataset, while the four datasets collected by \cite{crop_resize} are taken as the four unknown-class datasets respectively: ImageNet-crop, ImageNet-resize, LSUN-crop, and LSUN-resize, which are cropped or resized images from 200-class TingImageNet and 10-class LSUN \cite{LSUN}.

\subsubsection{Fine-Grained Datasets}

Under the cross-dataset setting, three semantic shift datasets which contain higher-resolution images from different subclasses of birds, cars, and aircrafts are used in \cite{IEEE-Access, good-closed-set, STAN_OSFGR}:
\begin{enumerate}
	\item[-] \textbf{CUB}: In this dataset, the images are taken from the Caltech-UCSD Birds (CUB) dataset (CUB-200-2011) \cite{CUB}, which contains 11788 200-class bird images with both labels and attribute tags, including 5994 training images and 5794 testing images with various image sizes. 100 classes are randomly selected as the known classes, while the rest 100 classes are further divided into three groups of unknown classes according to the attribute similarity between each unknown class and the whole known classes: `Easy' group that contains 32 classes more distinguished from the known classes, `Hard' group that contains 34 classes more similar to the known classes, and `Medium' group that contains the rest 34 classes. 
	
	\item[-] \textbf{FGVC-Aircraft}: In this dataset, the images are taken from the FGVC-Aircraft-2013b dataset \cite{Aircraft}, which likewise contains 10000 100-class car images with both labels and attribute tags, including 6667 training images and 3333 testing images with different image sizes. 50 classes are randomly selected as the known classes, while the rest 50 classes are also divided into three difficulty groups similar to CUB: 20-class `Easy' group, 13-class `Hard' group, and 17-class `Medium' group.
	
	\item[-] \textbf{Stanford-Cars}: In this dataset, the images are taken from the Stanford-Cars dataset \cite{Stanford-Cars}, which contains 16185 196-class aircraft images with labels, including 8144 training images and 8041 testing images (360$\times$240). In \cite{STAN_OSFGR}, the first 98 classes are selected as the known classes, while the rest 98 classes are taken as the unknown classes.
		
\end{enumerate}

Under the cross-dataset setting, the subset of FGVC-Aircraft that contains above selected 50 known classes is taken as the known-class dataset, while the 200-class CUB and 196-class Stanford-Cars datasets are taken as the two unknown-class datasets respectively: Aircraft-CUB, Aircraft-Stanford-Cars. 

\begin{table*}[t]\small
	\centering  
	\caption{ACC results on coarse-grained datasets under the standard-dataset setting.}  
	\centerline{
		\begin{tabular}{m{0.4cm}<{\raggedright}m{2.8cm}<{\centering}m{3.2cm}<{\centering}m{1.6cm}<{\centering}m{0.8cm}<{\centering}m{0.8cm}<{\centering}m{1cm}<{\centering}m{1.8cm}<{\centering}m{1.6cm}<{\centering}}
			\toprule
			Year & Method & Backbone & Architecture & MNIST & SVHN & CIFAR10 & CIFAR+10/+50 & TinyImageNet  \\
			\midrule
			2016 & OpenMax~\cite{OpenMax} & Convs* & E & 0.995 & 0.947 & 0.801 & - & -  \\
			2017 & G-OpenMax~\cite{G-openmax} & Convs* & G+D* & 0.996 & 0.948 & 0.816 & - & -  \\
			2018 & OSRCI~\cite{OSRCI} & Convs* & E+D+G & 0.996 & 0.951 & 0.821 & - & -  \\
			2019 & CROSR~\cite{CROSR} & Convs* & E+D & 0.992 & 0.945 & 0.930 & - & -  \\
			2019 & C2AE~\cite{C2AE} & Convs* & E+D & 0.992 & 0.936 & 0.910 & 0.919 & 0.430 \\
			2020 & CPN~\cite{CPN} & Convs* & E & 0.997 & 0.967 & 0.929 & - & - \\
			2020 & GDFR~\cite{GDFR} & Convs* & E+D+E & - & 0.973 & 0.951 & 0.974 & 0.559 \\
			2020 & CGDL~\cite{CGDL} & Convs* & E+D & 0.996 & 0.942 & 0.912 & 0.914 & 0.445 \\
			2020 & Hybrid~\cite{OpenHybrid} & Convs* & E+D & 0.995 & 0.962 & 0.926 & 0.937 & 0.632  \\
			2021 & GCM-CF~\cite{GCM_CF} & Convs* & E+D & 0.995 & - & 0.873 & 0.917 & 0.329  \\
			2021 & PROSER~\cite{PROSER} & Convs* & E & - & 0.964 & 0.926 & - & 0.521  \\
			2021 & EGT~\cite{Geometric} & Convs* & E & - & 0.977 & 0.943 & 0.959 & 0.656  \\
			2021 & Capsule~\cite{Capsule} & ResNet* & E+D & 0.994 & 0.984 & 0.952 & 0.969 & 0.774  \\
			2021 & CAC~\cite{CAC} & Convs* & E & 0.998 & 0.970 & 0.934 & 0.952 & 0.759  \\
			2021 & ARPL+CS~\cite{ARPL} & ResNet* & E+G+D* & - & - & 0.940 & - & - \\
			2021 & DivAR~\cite{DivAR} & Wide-ResNet*+ResNet* & E+E & - & 0.973 & 0.964 & 0.973 & 0.795  \\
			2021 & OpenGAN~\cite{OpenGAN} & ResNet* & G+D* & - & - & - & 0.940 & 0.684  \\
			2022 & OpenAUC~\cite{OpenAUC} & Convs* & E & 0.995 & - & 0.876 & 0.920 & 0.339  \\
			2022 & OVRN-CD~\cite{OVRN_CD} & Convs*/ResNet* & E & 0.998 & 0.975 & 0.932 & - & -  \\
			2022 & DIAS~\cite{Difficulty} & Convs* & G+D* & 0.997 & 0.970 & 0.947 & 0.964 & 0.700  \\ 
			2022 & Class-inclusion~\cite{Background_class} & Wide-ResNet* & E & - & 0.974 & 0.973 & 0.976 & 0.802  \\
			2022 & CSSR~\cite{CSSR_OSR} & ResNet*/Wide-ResNet* & E+E+D & - & - & 0.953 & - & - \\
			2022 & PMAL~\cite{PMAL} & Wide-ResNet* & E & 0.998 & 0.971 & 0.975 & 0.981 & 0.847  \\
			2022 & Cross-Entropy+~\cite{good-closed-set} & ResNet* & E & - & - & - & 0.958 & 0.862  \\
			2022 & OPG~\cite{OPG_OSR} & Convs* & E & - & 0.960 & 0.950 & 0.960 & 0.720  \\
			2022 & MoEP-AE~\cite{MoEP-AE} & Transformer* & E+D & 0.998 & 0.984 & 0.987 & 0.983 & 0.965 \\
			2023 & MGPL~\cite{MGPL} & ResNet* & E+D & 0.996 & 0.967 & 0.932 & - & 0.547  \\
			2023 & SLCPL~\cite{SLCP} & Convs* & E & 0.998 & 0.971 & 0.946 & - & -  \\
			2023 & OpenMix+~\cite{OpenMix} & Convs* & E & 0.995 & - & 0.953 & 0.968 & 0.584  \\
			2023 & ODL~\cite{ODL} & Convs* & E & 0.998 & 0.969 & 0.931 & 0.957 & 0.735  \\
			2023 & AKPF++~\cite{AKPF} & Convs* & G+D*+E & 0.998 & 0.969 & 0.960 & - & -  \\
			2023 & ConOSR~\cite{ConOSR} & Convs* & E & - & - & 0.946 & - & 0.661  \\
			2023 & IT-OSR-TransP~\cite{IT_OSR} & Transformer* & E+G+D* & 0.997 & 0.980 & 0.988 & 0.988 & 0.945 \\
			2023 & HAN-OSFGR~\cite{STAN_OSFGR} & Transformer* & E & - & - & - & 0.988 & 0.953  \\
			\bottomrule
	\end{tabular}}
	\label{table: standard_dataset_coarse_grained_ACC}
\end{table*}

\begin{table*}[t]\small
	\centering  
	\caption{AUROC results on coarse-grained datasets under the standard-dataset setting.}  
	\centerline{
		\begin{tabular}{m{0.4cm}<{\raggedright}m{2.8cm}<{\centering}m{3.2cm}<{\centering}m{1.6cm}<{\centering}m{0.8cm}<{\centering}m{0.8cm}<{\centering}m{1cm}<{\centering}m{1.8cm}<{\centering}m{1.6cm}<{\centering}}
			\toprule
			Year & Method & Backbone & Architecture & MNIST & SVHN & CIFAR10 & CIFAR+10/+50 & TinyImageNet  \\
			\midrule
			2016 & OpenMax~\cite{OpenMax} & Convs* & E & 0.981 & 0.894 & 0.695 & 0.817/0.796 & 0.576  \\
			2017 & G-OpenMax~\cite{G-openmax} & Convs* & G+D* & 0.984 & 0.896 & 0.675 & - & 0.580  \\
			2018 & OSRCI~\cite{OSRCI} & Convs* & E+D+G & 0.988 & 0.910 & 0.699 & 0.838/0.827 & 0.586  \\
			2019 & CROSR~\cite{CROSR} & Convs* & E+D & 0.991 & 0.899 & - & - & 0.589  \\
			2019 & C2AE~\cite{C2AE} & Convs* & E+D & 0.989 & 0.922 & 0.895 & 0.955/0.937 & 0.748  \\
			2020 & CPN~\cite{CPN} & Convs* & E & 0.987 & 0.924 & 0.828 & 0.881/0.879 & 0.639  \\
			2020 & GDFR~\cite{GDFR} & Convs* & E+D+E & - & 0.955 & 0.831 & 0.915/0.913 & 0.647 \\
			2020 & CGDL~\cite{CGDL} & Convs* & E+D & 0.994 & 0.935 & 0.903 & 0.959/0.950 & 0.762  \\
			2020 & Hybrid~\cite{OpenHybrid} & Convs* & E+D & 0.995 & 0.947 & 0.950 & 0.962/0.955 & 0.793  \\
			2021 & GCM-CF~\cite{GCM_CF} & Convs* & E+D & 0.960 & - & 0.733 & 0.833/0.820 & 0.591  \\
			2021 & PROSER~\cite{PROSER} & Convs* & E & - & 0.943 & 0.891 & 0.960/0.953 & 0.693  \\
			2021 & EGT~\cite{Geometric} & Convs* & E & - & 0.958 & 0.821 & 0.937/0.930 & 0.709  \\
			2021 & Capsule~\cite{Capsule} & ResNet* & E+D & 0.992 & 0.956 & 0.835 & 0.888/0.889 & 0.715  \\
			2021 & CAC~\cite{CAC} & Convs* & E & 0.985 & 0.938 & 0.803 & 0.863/0.872 & 0.772  \\
			2021 & ARPL+CS~\cite{ARPL} & ResNet* & E+G+D* & 0.997 & 0.967 & 0.910 & 0.971/0.951 & 0.782  \\
			2021 & DivAR~\cite{DivAR} & Wide-ResNet*+ResNet* & E+E & - & 0.965 & 0.885 & 0.955/0.946 & 0.787  \\
			2021 & OpenGAN~\cite{OpenGAN} & ResNet* & G+D* & 0.999 & 0.988 & 0.973 & 0.981/0.983 & 0.907  \\
			2022 & OpenAUC~\cite{OpenAUC} & Convs* & E & 0.977 & - & 0.761 & 0.860/0.849 & 0.614  \\
			2022 & OVRN-CD~\cite{OVRN_CD} & Convs*/ResNet* & E & 0.989 & 0.941 & 0.903 & 0.907/0.902 & 0.730  \\
			2022 & DIAS~\cite{Difficulty} & Convs* & G+D* & 0.992 & 0.943 & 0.850 & 0.920/0.916 & 0.731  \\
			2022 & Class-inclusion~\cite{Background_class} & Wide-ResNet* & E & - & 0.956 & 0.948 & 0.961/0.957 & 0.785  \\
			2022 & CSSR~\cite{CSSR_OSR} & ResNet*/Wide-ResNet* & E+E+D & - & 0.979 & 0.913 & 0.963/0.962 & 0.823  \\
			2022 & PMAL~\cite{PMAL} & Wide-ResNet* & E & 0.997 & 0.970 & 0.951 & 0.978/0.969 & 0.831  \\
			2022 & Cross-Entropy+~\cite{good-closed-set} & ResNet* & E & 0.984 & 0.959 & 0.910 & 0.954/0.939 & 0.826  \\
			2022 & OPG~\cite{OPG_OSR} & Convs* & E & - & 0.890 & 0.831 & 0.962/0.961 & 0.880  \\
			2022 & MoEP-AE~\cite{MoEP-AE} & Transformer* & E+D & 0.996 & 0.969 & 0.962 & 0.976/0.975 & 0.952  \\
			2023 & MGPL~\cite{MGPL} & ResNet* & E+D & - & 0.957 & 0.840 & 0.927/0.918 & 0.730  \\
			2023 & SLCPL~\cite{SLCP} & Convs* & E & 0.994 & 0.952 & 0.861 & 0.916/0.888 & 0.749  \\
			2023 & OpenMix+~\cite{OpenMix} & Convs* & E & 0.981 & - & 0.869 & 0.931/0.925 & 0.751  \\
			2023 & ODL~\cite{ODL} & Convs* & E & 0.995 & 0.943 & 0.857 & 0.891/0.883 & 0.764  \\
			2023 & AKPF++~\cite{AKPF} & Convs* & G+D*+E & 0.997 & 0.968 & 0.916 & 0.973/0.954 & 0.797  \\
			2023 & ConOSR~\cite{ConOSR} & Convs* & E & 0.997 & 0.991 & 0.942 & 0.981/0.973 & 0.809  \\
			2023 & IT-OSR-TransP~\cite{IT_OSR} & Transformer* & E+D+G* & 0.999 & 0.983 & 0.965 & 0.991/0.993 & 0.943  \\
			2023 & HAN-OSFGR~\cite{STAN_OSFGR} & Transformer* & E & - & - & - & 0.994/0.991 & 0.945  \\
			\bottomrule
	\end{tabular}}
	\label{table: standard_dataset_coarse_grained_AUROC}
\end{table*}

\begin{table*}[t]\small
	\centering  
	\caption{Macro-F1 results on coarse-grained datasets under the cross-dataset setting.}  
	\centerline{
		\begin{tabular}{m{0.8cm}<{\raggedright}m{2.8cm}<{\centering}m{3.4cm}<{\centering}m{1.8cm}<{\centering}m{1.8cm}<{\centering}m{1.3cm}<{\centering}m{1.3cm}<{\centering}m{1.3cm}<{\centering}}
			\toprule
			Year & Method & Backbone & Architecture & ImageNet-crop & ImageNet-resize & LSUN-crop & LSUN-resize  \\
			\midrule
			2016 & OpenMax~\cite{OpenMax} & Convs* & E & 0.660 & 0.684 & 0.657 & 0.668  \\
			2018 & OSRCI~\cite{OSRCI} & Convs* & E+D+G & 0.636 & 0.635 & 0.650 & 0.648  \\
			2019 & CROSR~\cite{CROSR} & Convs* & E+D & 0.721 & 0.735 & 0.720 & 0.749  \\
			2019 & C2AE~\cite{C2AE} & Convs* & E+D & 0.837 & 0.826 & 0.783 & 0.801  \\
			2020 & CPN~\cite{CPN} & Convs* & E & 0.850 & 0.835 & 0.853 & 0.884  \\
			2020 & GDFR~\cite{GDFR} & Convs* & E+D+E & 0.757 & 0.792 & 0.751 & 0.805  \\
			2020 & CGDL~\cite{CGDL} & Convs* & E+D & 0.840 & 0.832 & 0.806 & 0.812  \\
			2020 & Hybrid~\cite{OpenHybrid} & Convs* & E+D & 0.802 & 0.786 & 0.790 & 0.757  \\
			2021 & GCM-CF~\cite{GCM_CF} & Convs* & E+D & 0.666 & 0.652 & 0.680 & 0.666  \\
			2021 & PROSER~\cite{PROSER} & Convs* & E & 0.849 & 0.824 & 0.867 & 0.856  \\
			2021 & EGT~\cite{Geometric} & Convs* & E & 0.829 & 0.794 & 0.826 & 0.803  \\
			2021 & Capsule~\cite{Capsule} & ResNet* & E+D & 0.857 & 0.834 & 0.868 & 0.882  \\
			2021 & CAC~\cite{CAC} & Convs* & E & 0.764 & 0.752 & 0.756 & 0.777  \\
			2021 & ARPL+CS~\cite{ARPL} & ResNet* & E+G+D* & 0.796 & 0.796 & 0.794 & 0.885  \\
			2022 & OpenAUC~\cite{OpenAUC} & Convs* & E & 0.679 & 0.696 & 0.689 & 0.688  \\
			2022 & OVRN-CD~\cite{OVRN_CD} & Convs*/ResNet* & E & 0.835 & 0.825 & 0.846 & 0.839  \\
			2022 & Class-inclusion~\cite{Background_class} & Wide-ResNet* & E & 0.876 & 0.869 & 0.880 & 0.877  \\
			2022 & CSSR~\cite{CSSR_OSR} & ResNet*/Wide-ResNet* & E+E+D & 0.929 & 0.909 & 0.941 & 0.935  \\
			2022 & MoEP-AE~\cite{MoEP-AE} & Transformer* & E+D & 0.906 & 0.883 & 0.925 & 0.899  \\
			2023 & MGPL~\cite{MGPL} & ResNet* & E+D & 0.862 & 0.862 & 0.869 & 0.868  \\
			2023 & SLCPL~\cite{SLCP} & Convs* & E & 0.867 & 0.859 & 0.865 & 0.892  \\
			2023 & OpenMix+~\cite{OpenMix} & Convs* & E & 0.865 & 0.887 & 0.878 & 0.899  \\
			2023 & ODL~\cite{ODL} & Convs* & E & 0.861 & 0.842 & 0.871 & 0.856  \\
			2023 & AKPF++~\cite{AKPF} & Convs* & G+D*+E & 0.867 & 0.860 & 0.867 & 0.898  \\
			2023 & ConOSR~\cite{ConOSR} & Convs* & E & 0.891 & 0.843 & 0.912 & 0.881  \\
			2023 & IT-OSR-TransP~\cite{IT_OSR} & Transformer* & E+D+G* & 0.971 & 0.959 & 0.973 & 0.971  \\
			\bottomrule
	\end{tabular}}
	\label{table: cross_dataset_coarse_grained_macro_F1}
\end{table*}

\begin{table*}\small
	\centering
	\caption{ACC, AUROC, and OSCR results on the fine-grained CUB dataset under the standard-dataset setting.}
	\begin{tabular}{m{0.8cm}<{\centering}m{2.8cm}<{\centering}m{1.8cm}<{\centering}m{1.6cm}<{\centering}m{0.8cm}<{\centering}m{0.9cm}<{\centering}m{0.9cm}<{\centering}m{0.9cm}<{\centering}m{0.9cm}<{\centering}m{0.9cm}<{\centering}m{0.9cm}<{\centering}}
		\toprule
		\multirow{2}{*}{Year} & \multirow{2}{*}{Method} & \multirow{2}{*}{Backbone} & \multirow{2}{*}{Architecture} & \multirow{2}{*}{ACC} & \multicolumn{3}{c}{AUROC} & \multicolumn{3}{c}{OSCR} \\
		\cmidrule{6-8}\cmidrule{9-11}
		& & & & & E & M+H & All & E & M+H & All  \\
		\midrule
		2016 & OpenMax \cite{OpenMax} & ResNet* & E & 0.913 & 0.929 & 0.854 & 0.878 & - & - & -  \\
		2018 & GCPL \cite{GCPL} & Convs* & E & 0.783 & 0.805 & 0.689 & 0.725 & 0.711 & 0.592 & 0.630  \\
		2020 & Hybrid \cite{OpenHybrid} & Convs* & E+D & 0.683 & 0.873 & 0.801 & 0.824 & 0.662 & 0.590 & 0.613  \\
		2021 & GMVAE-OSR \cite{GMVAE-OSR} & Convs* & E+D & 0.725 & 0.826 & 0.727 & 0.758 & 0.694 & 0.591 & 0.623  \\
		2021 & OpenGAN \cite{OpenGAN} & ResNet* & G+D* & 0.799 & 0.801 & 0.736 & 0.757 & 0.725 & 0.677 & 0.692  \\		
		2021 & CAMV \cite{IEEE-Access} & ResNet* & E & 0.836 & 0.845 & 0.756 & 0.784 & 0.746 & 0.678 & 0.699  \\
		2021 & ARPL+ \cite{ARPL} & ResNet* & E & 0.859 & 0.835 & 0.755 & 0.780 & 0.760 & 0.696 & 0.716  \\
		2021 & PROSER \cite{PROSER} & ResNet* & E & 0.913 & 0.925 & 0.838 & 0.865 & - & - & - \\
		2022 & Cross-Entropy+ \cite{good-closed-set} & ResNet* & E & 0.862 & 0.883 & 0.793 & 0.821 & 0.798 & 0.731 & 0.752  \\
		2022 & OpenAUC \cite{OpenAUC} & Convs* & E & 0.862 & 0.888 & 0.807 & 0.832 & - & - & - \\
		2022 & MoEP-AE \cite{MoEP-AE} & Transformer* & E+D & 0.948 & 0.957 & 0.852 & 0.885 & 0.915 & 0.822 & 0.851  \\
		2022 & Trans-AUG \cite{OSR_transformer2} & Transformer* & E & 0.950 & 0.953 & 0.850 & 0.883 & 0.914 & 0.825 & 0.853  \\
		2023 & ODL \cite{ODL} & Convs* & E & 0.866 & - & - & 0.831 & - & - & 0.762  \\
		2023 & HAN-OSFGR \cite{STAN_OSFGR} & Transformer* & E & 0.959 & 0.965 & 0.891 & 0.914 & 0.934 & 0.870 & 0.890  \\
		\bottomrule
	\end{tabular}
	\label{table: standard_dataset_fine_grained_CUB}
\end{table*}


\begin{table*}\small
	\centering
	\caption{ACC, AUROC, and OSCR results on the fine-grained FGVC-Aircraft and Stanford-Cars datasets under the standard-dataset setting.}
	\begin{tabular}{m{0.8cm}<{\centering}m{2.8cm}<{\centering}m{1.8cm}<{\centering}m{1.6cm}<{\centering}m{0.8cm}<{\centering}m{1.6cm}<{\centering}m{1.6cm}<{\centering}m{0.8cm}<{\centering}m{1cm}<{\centering}m{1cm}<{\centering}}
		\toprule
		\multirow{2}{*}{Year} & \multirow{2}{*}{Method} & \multirow{2}{*}{Backbone} & \multirow{2}{*}{Architecture} & \multicolumn{3}{c}{FGVC-Aircraft} & \multicolumn{3}{c}{Stanford-Cars} \\
		\cmidrule{5-7}\cmidrule{8-10}
		& & & & ACC & AUROC(All) & OSCR(All) & ACC & AUROC & OSCR  \\
		\midrule
		2016 & OpenMax \cite{OpenMax} & ResNet* & E & 0.867 & 0.853 & - & - & - & -  \\
		2018 & GCPL \cite{GCPL} & Convs* & E & 0.823 & 0.801 & 0.750 & 0.688 & 0.735 & 0.663  \\
		2020 & Hybrid \cite{OpenHybrid} & Convs* & E+D & 0.735 & 0.843 & 0.684 & 0.675 & 0.699 & 0.637  \\
		2021 & GMVAE-OSR \cite{GMVAE-OSR} & Convs* & E+D & 0.814 & 0.801 & 0.748 & 0.683 & 0.756 & 0.670  \\
		2021 & OpenGAN \cite{OpenGAN} & ResNet* & G+D* & 0.807 & 0.794 & 0.744 & 0.704 & 0.763 & 0.682  \\
		2021 & CAMV \cite{IEEE-Access} & ResNet* & E & 0.868 & 0.829 & 0.760 & 0.754 & 0.832 & 0.724  \\
		2021 & ARPL+ \cite{ARPL} & ResNet* & E & 0.915 & 0.814 & 0.783 & - & - & -  \\
		2021 & PROSER \cite{PROSER} & ResNet* & E & 0.860 & 0.845 & - & - & - & -  \\
		2022 & Cross-Entropy+ \cite{good-closed-set} & ResNet* & E & 0.917 & 0.858 & 0.826 & 0.768 & 0.859 & 0.735  \\
		2022 & MoEP-AE \cite{MoEP-AE} & Transformer* & E+D & 0.894 & 0.856 & 0.834 & 0.869 & 0.922 & 0.819  \\
		2022 & Trans-AUG \cite{OSR_transformer2} & Transformer* & E & 0.898 & 0.828 & 0.772 & 0.860 & 0.910 & 0.809  \\
		2023 & ODL \cite{ODL} & Convs* & E & 0.919 & 0.851 & 0.817 & - & - & -  \\
		2023 & HAN-OSFGR \cite{STAN_OSFGR} & Transformer* & E & 0.927 & 0.896 & 0.857 & 0.888 & 0.936 & 0.849  \\
		\bottomrule
	\end{tabular}
	\label{table: standard_dataset_fine_grained_Aircraft_Cars}
\end{table*}

\begin{table*}[t]\small
	\centering
	\caption{Macro-F1 results on the two fine-grained datasets Aircraft-CUB and Aircraft-Cars under the cross-dataset setting.}
	\begin{tabular}{m{1.8cm}<{\centering}m{3.5cm}<{\centering}m{2.5cm}<{\centering}m{2.2cm}<{\centering}m{2.5cm}<{\centering}m{2.5cm}<{\centering}}
		\toprule
		Year & Method & Backbone & Architecture & Aircraft-CUB & Aircraft-Cars \\
		\midrule
		2018 & GCPL \cite{GCPL} & Convs* & E & 0.435 & 0.812  \\
		2020 & Hybrid \cite{OpenHybrid} & Convs* & E+D & 0.473 & 0.840 \\
		2021 & GMVAE-OSR \cite{GMVAE-OSR} & Convs* & E+D & 0.454 & 0.826   \\
		2021 & OpenGAN \cite{OpenGAN} & ResNet* & G+D* & 0.426 & 0.819 \\	
		2021 & CAMV \cite{IEEE-Access} & ResNet* & E & 0.461 & 0.833 \\	
		2021 & ARPL+ \cite{ARPL} & ResNet* & E & 0.469 & 0.831 \\
		2022 & Cross-Entropy+ \cite{good-closed-set} & ResNet* & E & 0.482 & 0.869 \\
		2022 & MoEP-AE \cite{MoEP-AE} & Transformer* & E+D & 0.529 & 0.876 \\
		2022 & Trans-AUG \cite{OSR_transformer2} & Transformer* & E & 0.508 & 0.882 \\
		2023 & HAN-OSFGR \cite{STAN_OSFGR} & Transformer* & E & 0.562 & 0.911 \\
		\bottomrule
	\end{tabular}
	\label{table: cross_dataset_fine_grained}
\end{table*}

\subsection{Metrics}

Here, we introduce the commonly-used evaluation metrics in the OSR task. The OSR task aims to not only accurately accept and classify the multi-class known-class testing samples, but also correctly reject the unknown-class testing samples. To evaluate the model performance based on the above targets, ACC and AUROC are the two most commonly-used metrics under the standard-dataset setting on both coarse-grained and fine-grained images. Besides, as suggested in \cite{CGDL, CROSR, C2AE, GDFR, PROSER}, macro-F1 score is also adopted for measuring the multi-class open-set classification performance under the cross-dataset setting. As suggested in \cite{good-closed-set}, OSCR \cite{OSCR} is also adopted for simultaneously measuring the closed-set classification performance and the open-set rejection performance on the fine-grained datasets. The details of the four evaluation metrics are shown as follows:
\begin{enumerate}
	\item[-] \textbf{ACC}: Top-1 accuracy (ACC) is the commonly-used metric in the closed-set recognition task. In the OSR task, this metric only considers the know-class testing samples. It is calculated by the proportion of the correctly classified known-class testing samples to the whole known-class testing samples.
	
	\item[-] \textbf{AUROC}: The area under the receiver operating characteristic (ROC) curve (AUROC) is a threshold-independent metric. In the OSR task, this metric considers all known classes as one class, while all unknown classes as another class, and measures the binary classification performance at various threshold settings. AUROC shows to what extent the model can classify the two classes. The ROC curve is depicted by the false positive rate (FPR) as abscissa and the true positive rate (TPR) as vertical coordinate. TPR and FPR are respectively calculated by:
	\begin{align}
	TPR = \dfrac{TP}{TP+FN}  \\
	FPR = \dfrac{FP}{FP+TN}
	\end{align}
	where $TP$ and $FN$ denote the numbers of the known-class testing samples that are correctly accepted as known classes and wrongly rejected as unknown classes, $FP$ and $TN$ denote the numbers of the unknown-class testing samples that are wrongly accepted as known classes and correctly rejected as unknown classes.
	
	\item[-] \textbf{macro-F1}: The macro-F1 score is a threshold-dependent metric that measures multi-class classification performance. In the OSR task, this metric considers all unknown classes as an additional class to the $C$ known classes, \emph{i.e.}, the ($C+1$)-th class. It is calculated based on the average precision $P_{macro}$ and the average recall $R_{macro}$, which are respectively calculated by:
	\begin{align}
	P_{macro} = \dfrac{1}{C+1} \sum_{i=1}^{C+1} \dfrac{TP_i}{TP_i+FP_i}   \\
	R_{macro} = \dfrac{1}{C+1} \sum_{i=1}^{C+1} \dfrac{TP_i}{TP_i+FN_i} 
	\end{align}
	where $TP_i$, $TN_i$, $FP_i$, and $FN_i$ denote the true positives, true negatives, false positives, and false negatives of the $i$-th class ($i \in \{ 1,2,..., C+1 \}$), respectively. Hence, the macro-F1 score is calculated by:
	\begin{align}
	F_{1\_macro} = 2 \times \dfrac{P_{macro} \times R_{macro}}{P_{macro} + R_{macro}}  
	\end{align}
	
	\item[-] \textbf{OSCR}: The Open-Set Classification Rate (OSCR) \cite{OSCR} is also a threshold-independent metric that simultaneously measures the $C$-class classification performance on the known-class testing samples as well as the binary classification performance on distinguishing unknown classes from known classes. Similar to AUROC, it is the area under another curve, which is depicted by correct classification rate (CCR) as abscissa and the newly-defined FPR as vertical coordinate. Here, CCR denotes the proportion of the known-class testing samples that are correctly accepted as known classes as well as correctly classified, while the newly defined FPR denotes the proportion of the unknown-class testing samples that are wrongly accepted as known classes. A larger OSCR indicates better performance in not only accepting and classifying known-class samples but also rejecting unknown-class samples.
	
\end{enumerate}

\subsection{Comparison}

In this subsection, we provide the comparison results for some representative OSR methods both on coarse-grained datasets and on fine-grained datasets under the two above-mentioned dataset settings.

\subsubsection{Comparison on Coarse-Grained Datasets}

The ACC and AUROC results of 34 representative OSR methods that have been evaluated on the coarse-grained datasets under the standard-dataset setting are reported in Tables \ref{table: standard_dataset_coarse_grained_ACC} and \ref{table: standard_dataset_coarse_grained_AUROC}, where the results of the comparative methods are cited from their original papers or the papers citing them and sorted under both the year and the AUROC metric on TinyImageNet. Besides, the corresponding macro-F1 scores under the cross-dataset setting are reported in Table \ref{table: cross_dataset_coarse_grained_macro_F1}, where the results of some methods are absent since they are not reported anywhere. For maintaining the consistency, the macro-F1 results are also sorted following the above strategy. 

The backbones and the network architectures are also listed for better comparison. ``Convs", ``ResNet", ``Wide-ResNet", and ``Transformer" denote that the corresponding methods are developed based on plain CNNs, ResNets, Wide-ResNets, and vision transformers, respectively. Since different methods usually adopt different layer configurations, in spite of using the same group of backbones, we take ``*"  for covering different configurations. Besides, ``+" and ``/" in Backbone item denote that the method combines two networks as its backbone and adopts different backbones on different datasets, respectively. ``E", ``G", ``D", and ``D*" in Architecture item denote encoder, generator, decoder, and discriminator, respectively. 

%
%
%

\subsubsection{Comparison on Fine-Grained Datasets}

The ACC, AUROC, and OSCR results of some OSR methods that have been reported on CUB, FGVC-Aircraft, and Stanford-Cars, under the standard-dataset setting are reported in Tables \ref{table: standard_dataset_fine_grained_CUB} and \ref{table: standard_dataset_fine_grained_Aircraft_Cars}. Besides, the corresponding macro-F1 scores under the cross-dataset setting are reported in Table \ref{table: cross_dataset_fine_grained}. The results in these tables are sorted by both the year and the OSCR metric on CUB. ``E", ``M", and ``H" denote the `Easy', `Medium', and `Hard' difficulty levels, respectively. As done in \cite{good-closed-set}, the `Medium' and `Hard' groups are reported jointly rather than separately on CUB. 

\subsubsection{Summary and Analysis on Comparative Results}

In summary, according to the above comparisons, 8 points can be seen from these tables:
\begin{enumerate}
	\item[-] As seen from Tables \ref{table: standard_dataset_coarse_grained_ACC} and \ref{table: standard_dataset_coarse_grained_AUROC}, most of the comparative methods both with more powerful backbones and even with simpler architectures outperform the methods with more lightweight backbones in most cases, demonstrating that a powerful backbone is helpful for boosting the model discriminability. Although a more powerful backbone generally leads to higher model complexity, which limits the architecture complexity, it can still combine with some elaborate-designed modules such as attention mechanisms in HAN-OSFGR \cite{STAN_OSFGR}, comprehensive data augmentation in Cross-Entropy+ \cite{good-closed-set}, lightweight reconstruction in CSSR \cite{CSSR_OSR} and MoEP-AE \cite{MoEP-AE}, prototype constraints in PMAL \cite{PMAL}, outlier exposure in Class-inclusion \cite{Background_class}, feature generation in OpenGAN \cite{OpenGAN} and IT-OSR-TransP \cite{IT_OSR}, etc., for further improving the model performance.
	
	\item[-] As seen from Tables \ref{table: standard_dataset_coarse_grained_ACC} and \ref{table: standard_dataset_coarse_grained_AUROC}, if a method yields a higher ACC, it generally also achieves a higher AUROC. In other words, in most of the comparative methods, a better closed-set classification performance roughly brings with a better open-set detection performance, which is consistent with the observation in \cite{good-closed-set}. However, there are also lots of counter-examples when comparing in pairs, where a significant higher AUROC corresponds to a similar or even significantly lower ACC and a better ACC corresponds to a poorer AUROC. This is mainly because that some unknown-class samples are similar to some known-class samples, separating these unknown-class samples from known classes would harm the known-class classification accuracy. Hence, How to achieve a better balance between open-set detection performance and closed-set classification performance is an open issue in the OSR community.
		
	\item[-] As seen from Tables \ref{table: standard_dataset_coarse_grained_ACC}-\ref{table: cross_dataset_coarse_grained_macro_F1}, the results under the cross-dataset setting are generally lower than those under the standard-dataset setting, mainly due to two reasons: (i) there also exists covariate shift between the training samples and unknown-class testing samples in addition to the semantic shift, which may cause the class confusion more easily. (ii) the macro-F1 metric simultaneously considers the closed-set classification performance and open-set detection performance, a model needs to not only correctly classify known-class samples, but also identify them as known classes based on a threshold.
		
	\item[-] As seen from Tables \ref{table: standard_dataset_coarse_grained_ACC}-\ref{table: cross_dataset_coarse_grained_macro_F1}, the recent OSR methods have achieved nearly saturated performance, \emph{i.e.}, approaching or higher than 95\%, especially with the transformer backbones. Hence, it is necessary to conduct evaluations on larger-scale, higher-resolution, and more difficult datasets.
	
	\item[-] As seen from Table \ref{table: standard_dataset_fine_grained_CUB}, the results in the `Medium' and `Hard' groups are generally lower than those in the `Easy' group, mainly because that the unknown-class images in harder groups usually have similar appearance to known-class images and they are only different in some fine-grained properties. Besides, the recent model performances also achieve or approach saturation in the `Easy' group, and identifying more difficult unknown-class samples mainly affects the overall performances. Hence, how to distinguish and make use of the harder unknown-class samples becomes a key to improving the model discriminability on fine-grained datasets. More effective acquisition and detailed processing of fine-grained information maybe helpful for distinguishing the samples in harder classes, which can then be utilized for model training based on some unsupervised techniques. 
	
	\item[-] As seen from Tables \ref{table: standard_dataset_fine_grained_CUB} and \ref{table: standard_dataset_fine_grained_Aircraft_Cars}, the methods with ResNet* backbones usually perform better than those with Convs* backbones, and the methods with Transformer* backbones usually perform better than those with ResNet* backbones. This is mainly because that the residual connections in ResNets deepen the network thus enhancing the network's ability to learn more complex features, and the multiple self-attention operations in transformers are helpful for capturing more fine-grained attentions on the semantic objects in images.
	
	\item[-] As seen from Table \ref{table: cross_dataset_fine_grained}, the results on Aircraft-CUB are significantly lower than those on Aircraft-Cars. This is mainly because that the distribution shift between FGVC-Aircraft and CUB is larger than that between FGVC-Aircraft and Stanford-Cars. Combining with the previous observations, we can find that both excessively larger distribution shift (\emph{e.g.}, the distribution between unknown-class samples on CUB and known-class samples on FGVC-Aircraft) and excessively smaller distribution shift (\emph{e.g.}, minor fine-grained differences between `Medium+Hard' group of unknown-class samples and known-class samples) would cause the class confusion. Since the distribution shift problem is an inherent issue in the OSR community, transductive learning may be helpful for alleviating the distribution shift.
	
	\item[-] As seen from Tables \ref{table: standard_dataset_coarse_grained_ACC}-\ref{table: cross_dataset_fine_grained}, generative OSR methods generally perform better than discriminator OSR methods under the standard-dataset setting and on coarse-grained datasets, mainly because that generative models are learned based on not only the relationships between inputs and outputs, but also the internal distribution structure of the data. However, such phenomenon is difficult to be observed either under cross-dataset setting or on fine-grained datasets, mainly because that the learned semantic-based distribution couldn't adapt to the covariate shift under the cross-dataset setting, and the distributions learned from similar samples would also be similar on fine-grained datasets. Hence, the discriminative methods that adopt various strategies for enhancing the feature discriminability would be more effective either under the cross-dataset setting or on the fine-grained datasets.
	
\end{enumerate}

%
%
%

\section{Open Issues and Future Research Directions}    \label{section: future}

In this section, we present some open issues in the OSR methods. Besides, we provide some future research directions for handling the OSR task from a new perspective, which may inspire the following works in the OSR and similar tasks.

\subsection{Open Issues}

Here, we present some open issues in the OSR task. 

\subsubsection{The Semantic Shift Problem}

The inherent issue in the OSR task is the semantic shift problem, where some new-class images would be encountered in the testing set. Since the deep learning models are data-driven models, only training the models based on known classes would make the models prefer the known classes, \emph{i.e.}, the models would wrongly predict the unknown-class samples as one of the known classes. Most of existing OSR methods are inductive methods, which assume that only known-class samples are available in the model training. Although they pursue more discriminative representations for known classes, hoping that unknown-class samples could be identified based on their deviation from known classes, the bias of the decision boundary between known classes and unknown classes still exists due to the absence of the real unknown classes. Transductive learning, however, has demonstrated its effectiveness in both other tasks and the OSR task. Up to now, there only exist two transductive OSR methods. Besides, some open sub-problems as mentioned in Sec. II.B also exist in the existing transductive OSR methods. Hence, how to effectively make use of the unlabeled testing samples is still worth studying, extremely the testing samples which are difficult to be identified.

\subsubsection{The Consistency Problem Between Classifying Known Classes and Identifying Unknown Classes}

The OSR task aims to simultaneously classify known-class samples as well as identify unknown-class samples. A good OSR model is desired to produce both high closed-set classification accuracy and high discrepancy between known classes and unknown classes. However, the two goals maybe sometimes inconsistent, when some hard unknown-class samples are confused with known-class samples in the feature space. In this case, separating these difficult unknown-class samples from the known-class ones is likely to harm the discriminability among different known classes. Such phenomenon can be also observed in the closed-set recognition task, where the accurate discrepancy among a part of the classes may lead to the confusion among some other classes. Hence, another open issue is raised: How to achieve consistently better performance or achieve a better balance between known-class classification and unknown-class identification? Mining and making use of the confused samples that are difficult to be identified may be a way to address this issue.

\subsubsection{Threshold for Distinguishing Between Known Classes and Unknown Classes}

Since the unknown-class samples are generally unavailable in the model training, most of existing OSR methods firstly train a $C$-way classifier and identify the unknown-class samples by comparing the $C$-class based identification score with a threshold. Hence, the choice of the threshold become essential to the open-set recognition performance. Most of existing OSR methods empirically choose a threshold for identification, which may not be suitable for all classes in all datasets. Zhou \emph{et al.} \cite{PROSER} proposed a class-specific threshold by a learnable strategy, which has paved a way for further researches about the threshold.

\subsection{Future Research Directions}

Here, we provide some future research directions for facilitating future works in handling the OSR task.

\subsubsection{Human Brain Mechanism Inspired Open-Set Recognition}

In the neuroscience field, human brains and animal brains are proved to have the ability of quickly recognizing new categories \cite{brain1, brain2, brain3}. Some recognition mechanisms can inspire the future OSR methods. Some existing OSR methods have already provided examples. Yang \emph{et al.} \cite{CPN} designed the class-specific feature prototypes inspired by the abstract memories of different classes in human brains. Huang \emph{et al.} \cite{MSD_Net} enabled the model to produce different reaction time for different images, aiming to improve the consistency with human behavior. Sun \emph{et al.} \cite{STAN_OSFGR} temporally aggregated the hierarchical attentioned features, inspired by the temporal attention mechanism in brains. Hence, it's a promising future research direction to refer from recognition mechanisms in brains.

\subsubsection{Multi-Modal Large Model Guided Open-Set Recognition}

With the rapid expansion of data volume and the improvement of hardware performance, deep neural networks are entering the era of multi-modal large models. Recently, a lot of multi-modal large models pretrained on multi-modal large-scale datasets have demonstrated their generalization ability in assisting many visual tasks \cite{MLLM_survey1, MLLM1, MLLM2, MLLM3, MLLM4, MLLM5}, such as few-shot and zero-shot image recognition tasks. The pretrained large models store abundant prior information about the open world, which is a promising auxiliary tool for handling the OSR task. A simple approach to make use of the large models is to tune their prompts, which provide context or parameter information about the input data for helping the large models to better understand and handle the specific tasks. 

Encouraged by the impressive generation ability of the pretrained large models, Qu \emph{et al.} \cite{MLLM_OSR1} collaborated several large models (ChatGPT \cite{ChatGPT}, DALL-E \cite{DALL-E}, CLIP \cite{CLIP}, and DINO \cite{DINO}) to make use of the rich implicit knowledge in a training-free manner to reduce the reliance on spurious-discriminative features. Their method was operated at two stages. At one stage, virtual unknown-class images and simulated names were generated by ChatGPT and DALL-E. At the other stage, a testing image was inferred based on two alignments with the generated images and the expanded list of both known classes and virtual unknown classes by CLIP and DINO. Although this method was intuitive and didn't need training, it was effective in handling the OSR task. To further make use of large models, Liao \emph{et al.} \cite{MLLM_OSR2} combined the open words and prompt tuning on large models for handling the OSR task. Different from \cite{MLLM_OSR1}, the open words were taken from WordNet \cite{WordNet} rather than generated by asking ChatGPT, and the learnable prompts improved the model adaptability to the downstream tasks. Besides, to XXX on larger-scale datasets, they firstly performed multiple independent group-wise prompt tuning on groups of fewer classes, and then predicted based on optimal sub-prompt. Furthermore, they proposed new baselines for fair comparison with large-model based OSR methods, which paved a way for future OSR methods based on prompt-tuning. How to make better use of the open-world pretrained large models based on new prompts and tuning strategies for handling the OSR task is still worth studying.

\section{Conclusion}   \label{section: conclusion}

In this paper, we present a comprehensive survey of open-set image recognition. Firstly, we propose a systematic taxonomy and review the existing DNN-based methods. Besides, we compare and analyze typical and state-of-the-art OSR methods on multiple datasets and under the two dataset deployments. Furthermore, we discuss some open issues and future directions in this community.


\ifCLASSOPTIONcaptionsoff
  \newpage
\fi


\small
\bibliographystyle{IEEEtran}
\bibliography{egbib_20221111}


\end{document}